\newcount\Comments  %
\documentclass[twoside]{article}
\usepackage[accepted]{aistats2016}

\usepackage{color}
\definecolor{darkgreen}{rgb}{0,0.5,0}
\definecolor{purple}{rgb}{1,0,1}
\newcommand{\kibitz}[2]{\ifnum\Comments=1\textcolor{#1}{#2}\fi}
\usepackage{rotating}
\usepackage[numbers]{natbib}
\usepackage{hyperref}
\usepackage{url}
\usepackage{subcaption}
\usepackage{hyperref}
\usepackage{graphicx} %
\usepackage{booktabs} 
\usepackage{amssymb}
\usepackage{times}
\usepackage{epsfig}
\usepackage{graphicx}
\usepackage{color}
\usepackage{url}
\usepackage{hyperref}
\usepackage{bbm}
\usepackage{multicol}

\providecommand{\dif}{\mathop{}\!\mathrm d}
\providecommand{\Ex}{\mathbb E}
\providecommand{\hide}[1]{}

\usepackage{amsmath,amsfonts}
\usepackage{amsopn,amssymb,amsthm,thmtools,thm-restate}

\theoremstyle{plain}
\newtheorem{thm}{Theorem}[section]
\newtheorem{lem}[thm]{Lemma}
\newtheorem{prop}[thm]{Proposition}
\newtheorem{cor}[thm]{Corollary}

\theoremstyle{definition}

\theoremstyle{remark}

\usepackage{bm} %
\usepackage{algorithm}%
\usepackage{algpseudocode}%
\newcommand{\vct}[1]{\boldsymbol{#1}} %
\newcommand{\mat}[1]{\boldsymbol{#1}} %
\newcommand{\field}[1]{\mathbb{#1}}
\newcommand{\R}{\field{R}} %
\newcommand{\T}{^{\textrm T}} %

\newcommand{\ProbOpr}[1]{\mathbb{#1}}
\newcommand{\expect}[2]{%
\ifthenelse{\equal{#2}{}}{\ProbOpr{E}_{#1}}
{\ifthenelse{\equal{#1}{}}{\ProbOpr{E}\left[#2\right]}{\ProbOpr{E}_{#1}\left[#2\right]}}} %
\newcommand{\var}[2]{%
\ifthenelse{\equal{#2}{}}{\ProbOpr{VAR}_{#1}}
{\ifthenelse{\equal{#1}{}}{\ProbOpr{VAR}\left[#2\right]}{\ProbOpr{VAR}_{#1}\left[#2\right]}}} %
\DeclareMathOperator{\Cov}{Cov}

\DeclareMathOperator{\argmax}{arg\,max}
\DeclareMathOperator{\argmin}{arg\,min}

\newcommand{\vmu}{\vct{\mu}}
\newcommand{\vgamma}{\vct{\gamma}}

\newcommand{\vx}{{\vct{x}}}
\newcommand{\vy}{\vct{y}}

\newcommand{\vu}{\vct{u}}

\newcommand{\va}{\vct{a}}
\newcommand{\vb}{\vct{b}}

\newcommand{\vf}{\vct{f}}

\newcommand{\vk}{\vct{k}}

\newcommand{\vv}{\vct{v}}

\newcommand{\mI}{\mat{I}}
\newcommand{\mK}{\mat{K}}

\newcommand{\mSigma}{\mat{\Sigma}}

\newcommand{\rt}{\tilde{r}} %
\newcommand{\bt}{\nu}

\newcommand{\EST}{{\sc{EST}}}
\newcommand{\ESTa}{{\sc{EST}}a}
\newcommand{\ESTn}{{\sc{EST}}n}
\newcommand{\EI}{{\sc{EI}}}
\newcommand{\PI}{{\sc{PI}}}
\newcommand{\RND}{{\sc{Rand}}}
\newcommand{\UCB}{{\sc{UCB}}}

\algnewcommand{\algorithmicgoto}{\textbf{go to}}%
\algnewcommand{\Goto}[1]{\algorithmicgoto~\ref{#1}}%
\begin{document}
\twocolumn[

\aistatstitle{Optimization as Estimation with Gaussian Processes in Bandit Settings}

\aistatsauthor{ Zi Wang \And Bolei Zhou \And Stefanie Jegelka }

\aistatsaddress{ MIT CSAIL \And MIT CSAIL \And MIT CSAIL } ]

\begin{abstract}
Recently, there has been rising interest in Bayesian optimization -- the optimization of an unknown function with assumptions usually expressed by a Gaussian Process (GP) prior. We study an optimization strategy that directly uses an estimate of the argmax of the function. This strategy offers both practical and theoretical advantages: no tradeoff parameter needs to be selected, and, moreover, we establish close connections to the popular GP-UCB and GP-PI strategies. Our approach can be understood as automatically and adaptively trading off exploration and exploitation in GP-UCB and GP-PI. We illustrate the effects of this adaptive tuning via bounds on the regret as well as an extensive empirical evaluation on robotics and vision tasks, demonstrating the robustness of this strategy for a range of performance criteria. %

\end{abstract} 
\section{Introduction}
\label{sec:intro}

The optimization of an unknown function that is expensive to evaluate is an important problem in many areas of science and engineering. Bayesian optimization uses probabilistic methods to address this problem. In particular, an increasingly popular direction has been to model smoothness assumptions on the function via a Gaussian Process (GP). The Bayesian approach provides a posterior distribution of the unknown function, and thereby uncertainty estimates that help decide where to evaluate the function next, in search of a maximum. Recent successful applications of this \emph{Bayesian optimization} framework include the tuning of hyperparameters for complex models and algorithms in machine learning, robotics, and computer vision~\cite{brochu2009,calandra2014experimental,krause2011contextual,lizotte2007,snoek2012practical,thornton13}. %

Despite progress on theory and applications of Bayesian optimization methods, the practitioner continues to face many options: there is a menu of algorithms, and their relations and tradeoffs are only partially understood. 
Typically, the points where the function is evaluated are selected sequentially; and the choice of the next point is based on observed function values at the previous points. %
Popular algorithms vary in their strategies to pick the next point: they select the point that maximizes the \emph{probability of improvement} (GP-PI)~\cite{kushner1964}; the \emph{expected improvement} (GP-EI)~\cite{mockus1974}; or an \emph{upper confidence bound} (GP-UCB)~\cite{srinivas2009gaussian} on the maximum function value. Another alternative is \emph{entropy search} (ES)~\cite{hennig2012}, which aims to minimize the uncertainty about the location of the optimum of the function. Each algorithm reduces the black-box function optimization problem to a series of optimization problems of known  \emph{acquisition functions}.

The motivations and analyses (if available) differ too: objectives include \emph{cumulative regret}, where every evaluation results in a reward or cost and the average of all function evaluations is compared to %
the maximum value of the function; %
 \emph{simple regret} that takes into account only the best value found so far~\cite{bubeck2009pure}; the performance under a fixed finite budget~\cite{grunewalder2010regret}; or the uncertainty about the location of the %
function maximizer~\cite{hennig2012}. Here, we focus on the established objectives of cumulative regret in bandit games.

Notably, many of the above algorithms involve tuning a parameter to trade off exploration and exploitation, and this can pose difficulties in practice~\cite{hennig2012}. Theoretical analyses help in finding good  parameter settings, but may be conservative in practice~\cite{srinivas2009gaussian}. %
Computing the \emph{acquisition function} and optimizing it to find the next point can be computationally very costly too. For example, the computation to decide which next point to evaluate for entropy search methods tends to be very expensive, while GP-PI, GP-EI and GP-UCB are much cheaper.

In this paper, we study an intuitive strategy that offers a compromise between a number of these approaches and, at the same time, establishes connections between them that help understand when theoretical results can be transferred. Our strategy uses the Gaussian Process to obtain an %
estimate of the argument that maximizes the unknown function $f$. %
The next point to evaluate is determined by this %
 estimate. This point, it turns out, is not necessarily the same as the the argument with the highest upper confidence bound.

This strategy has both practical and theoretical advantages. On the theoretical side, we show connections to the popular GP-UCB and GP-PI strategies, implying %
an intuitive and provably correct way of setting the parameters %
in those important methods. Moreover, we establish bounds on the regret of %
our estimation strategy. From a practical viewpoint, %
our strategy obviates any costly parameter tuning. In fact, we show that it corresponds to automatically and \emph{adaptively} tuning the parameters of GP-UCB and GP-PI. %
Our %
empirical evaluation includes problems from non-convex optimization, robotics, and computer vision. The experiments show that %
our strategy performs similarly to or even better than the best competitors in terms of cumulative regret. Although not designed to minimize simple regret directly, in practice our method also works well as measured by simple regret, or by the number of steps to reach a fixed regret value. Together, these  results suggest that our strategy is easy to use and empirically performs well across a spectrum of settings. %

\paragraph{Related work.} The practical benefits of Bayesian optimization have been shown in a number of applications~\cite{brochu2009,calandra2014experimental,lizotte2007,snoek2012practical,thornton13,wang2013}. 
Different Bayesian optimization algorithms differ in the selection criteria of the next point to evaluate, i.e., the acquisition function.
Popular criteria include the expected improvement (GP-EI)~\cite{mockus1974}, the probability of improving over a given threshold (GP-PI)~\cite{kushner1964}, and GP-UCB~\cite{srinivas2009gaussian}, which is motivated by upper confidence bounds for multi-armed bandit problems~\cite{auer2002finite,auer2002b}. GP-EI, GP-PI and GP-UCB have a parameter to select, and the latter two are known to be sensitive to this choice. Entropy search (ES)~\cite{hennig2012} and the related predictive entropy search (PES)~\cite{hernandez2014predictive} do not aim to minimize regret directly, but to maximize the amount of information gained about the optimal point. %
High-dimensional settings were considered in~\cite{djolonga2013high,wang2013}.
Extensive empirical comparisons include~\cite{calandra2014experimental,snoek2012practical,lizotte2012}.
Theoretical bounds on different forms of regret were established for GP-UCB~\cite{srinivas2009gaussian} and GP-EI~\cite{bull2011}. Other theoretical studies focus on simple regret~\cite{bubeck2009pure,grunewalder2010regret} or finite budgets~\cite{grunewalder2010regret}. In this work, in contrast, we are motivated by practical considerations.

\subsection{Background and Notation}
\label{sec:background}
Let $f(\cdot)\sim GP(0,k)$ be an unknown function we aim to optimize over a candidate set $\mathfrak X$. %
At time step $t$, we select point $\vx_t$ and observe a possibly noisy function evaluation $y_t=f(\vx_t)+\epsilon_t$, where $\epsilon_t$ are i.i.d.\ Gaussian noise $\mathcal N(0,\sigma^2)$.
Given the observations $\mathfrak D_t=\{(\vx_\tau,y_\tau)\}_{\tau=1}^{t}$ up to time $t$, we obtain the posterior mean and covariance of the function via the kernel matrix $\mK_t =\left[k(\vx_i,\vx_j)\right]_{\vx_i,\vx_j\in \mathfrak D_t}$ and $\vk_t(x) = [k(\vx_i,\vx)]_{\vx_i\in \mathfrak D_t}$~\cite{rasmussen2006gaussian}:
$\mu_{t}(\vx) = \vk_t(\vx)\T(\mK_t+\sigma^2\mI)^{-1}\vy_t$, and %
$k_{t}(\vx,\vx') = k(\vx,\vx') - \vk_t(\vx)\T(\mK_t+\sigma^2\mI)^{-1} \vk_t(\vx')$.
The posterior variance is given by $\sigma^2_{t}(\vx) = k_t(\vx,\vx)$. Furthermore, we denote by $Q(\cdot)$ the tail probability of the standard normal distribution $\phi(\cdot)$, and by $\Phi(\cdot)$ its cumulative probability. %

The Bayesian Optimization setting corresponds to a bandit game where, in each round $t$,
the player chooses a point $\vx_{t}$ and then observes $y_{t} = f(\vx_t) + \epsilon_t$. The \emph{regret} for round $t$ is defined as $\rt_t = \max_{\vx\in \mathfrak X}f(\vx) - f(\vx_{t})$. The \emph{simple regret} for any $T$ rounds is $r_T = \min_{t\in[1,T]} \rt_t$, and the (average) cumulative regret is $R_T =\frac{1}{T}\sum_{t=1}^T \rt_t$.

\subsection{Existing methods for GP optimization}
\label{sssec:existing}
We focus on the following three approaches for comparison, since they are most widely used in bandit settings.

\textbf{GP-UCB.}~\citet{srinivas2009gaussian} provide a detailed analysis for using upper confidence bounds~\cite{auer2002finite} with GP bandits. They propose the strategy %
$\vx_t = \argmax_{\vx\in \mathfrak X}\mu_{t-1}(\vx) + \lambda_t\sigma_{t-1}(\vx)$ where $\lambda_t = (2\log(|\mathfrak X|\pi^2 t^2/(6\delta)))^{\frac12}$ for finite $\mathfrak X$. Their regret bound holds with probability $1-\delta$. %

\textbf{GP-EI.} The GP-EI  strategy~\cite{mockus1974} selects the point maximizing the expected improvement over a pre-specified threshold $\theta_t$~\cite{mockus1974}. For GPs, this improvement is given in closed form as $\mathrm{EI}(x) = \mathbb{E}[ (f(x) - \theta_t)_+] = \left[\phi(\vgamma(\vx)) - \vgamma(\vx) Q(\vgamma(\vx))\right]\sigma_{t-1}(\vx)$, where $\vgamma(\vx) = \frac{\theta_t - \mu_{t-1}(\vx)}{\sigma_{t-1}(\vx)}$. %
A popular choice for the threshold is $\theta_t = \max_{\tau\in[1,t-1]}y_\tau$. 

\textbf{GP-PI.} The third strategy maximizes the probability $\mathrm{PI}(x) = \Pr[f(x) > \theta_t] = 1 - \Phi(\vgamma(\vx))$ of improving over a threshold $\theta_t$~\cite{kushner1964}, i.e., $\vx_t = \argmin_{\vx\in\mathfrak X} \vgamma(\vx)$. GP-PI is sensitive to the choice of $\theta_t$: as we will see in Section~\ref{sec:method}, $\theta_t$ trades off exploration and exploitation, and setting $\theta_t$ too low (e.g., $\theta_t = \max_{\tau\in[1,t-1]}y_\tau$) can result in getting stuck at a fairly suboptimal point. A popular choice in practice is $\theta_t =  \max_{\tau\in[1,t-1]}y_\tau + \epsilon$, for a chosen constant $\epsilon$.

\hide{
At time step $t$, GP-UCB chooses

$$\vx_r(t) = \argmax_{\vx\in \mathcal A} \mu_{t-1} (\vx) + \beta_{t}^{\frac12}\sigma_{t-1}(\vx)$$ 

to be the next point to evaluate. $\beta_{t}$ is the magic number that we have to set in order to balance \emph{exploration} and \emph{exploitation}. In the original paper~\cite{srinivas2009gaussian}, the authors proposed $\beta_t= 2\log(|\mathcal A|t^2\pi^2/6\gamma)$ where $f$ is sampled from a known GP prior, and $\gamma \in (0,1)$, $1-\gamma$ is a confidence level. 

They gave theoretical guarantees that the average of cumulative regret is 0 when $t$ goes to infinity. However in their experiments, they suggested that scaling $\beta_t$ down by a factor of 5 would give better results, without justifying the reason.%

In particular, if we set $\beta_t$ to be large, the algorithm will definitely explore too much before recognizing the optimizer, while if we set $\beta_t$ to be small, the algorithm will start to exploit too soon and as a result, it never pick the optimizer. %

GP-UCB will not evaluate the $\argmax$ unless its upper bound is higher than the upper bounds for all the other data points.
}

\section{Optimization as estimation}\label{sec:method}

In this work, we study an alternative criterion that provides an easy-to-use and tuning-free approach: we use the GP to estimate the $\argmax$ of $f$. %
In Section~\ref{subsec:connections}, we will see how, as a side effect, this criterion establishes connections between the above criteria. Our strategy eventually leads to tighter bounds than GP-UCB as shown in Section~\ref{sec:bounds}. %

Consider the posterior probability (in round $t$) that a fixed $\vx\in \mathfrak X$ is an $\argmax$ of $f$. We call this event $M_{\vx}$ and, for notational simplicity, omit the subscripts $t-1$ here.
The event $M_\vx$ is equivalent to the event that for all $\vx'\in\mathfrak X$, we have $v(\vx') := f(\vx')-f(\vx) \leq 0$. The difference $v(\vx')$ between two Gaussian variables is Gaussian: $v(\vx') \sim \mathcal{N}(\mu(\vx') - \mu(\vx), \sigma(\vx)^2+\sigma(\vx')^2-2k(\vx,\vx')^2)$. The covariance for any $\vx',\vx''\in \mathfrak X$ is $\Cov(v(\vx'),v(\vx'')) %
= \sigma(\vx)^2+k(\vx',\vx'')^2-k(\vx,\vx')^2-k(\vx,\vx'')^2$.

The random variables $\{v(\vx')\}_{\vx'\in\mathfrak X}$ determine the cumulative probability 
\begin{align}\label{eq:probmax}
\Pr[M_\vx|\mathfrak D] = \Pr[\forall \vx'\in\mathfrak X, v(\vx')\leq0 | \mathfrak{D}].
\end{align}
This probability may be specified via limits as e.g.\ in 
\cite[App.A]{hennig2012}. Moreover, due to the assumed smoothness of $f$, %
 it is reasonable to work with a discrete approximation and restrict the set of candidate points to be finite for now (we discuss discretization further in Section~\ref{sec:disc}). So the quantity in Eqn.~\eqref{eq:probmax} is well-defined.
Since computing $\Pr[M_\vx|\mathfrak D]$ for large $|\mathfrak X|$ can be costly, we use a ``mean-field'' approach and approximate $\{f(\vx)\}_{\vx\in\mathfrak X}$ by independent Gaussian random variables with means $\mu(\vx)$ and variances $\sigma(\vx)^2$ for all $\vx\in \mathfrak X$. %
Given be the maximum value $m$ of $f$, the probability of the event $M_\vx|m,\mathfrak{D}$ amounts to %
\begin{align}
\Pr[M_\vx | m,\mathfrak D] %
& \approx Q\Big(\frac{m-\mu(\vx)}{\sigma(\vx)}\Big)\prod_{\vx'\neq \vx}\Phi\Big(\frac{m-\mu(\vx')}{\sigma(\vx')}\Big). \nonumber
\end{align}
Our estimation strategy (EST) chooses to evaluate $\argmax_{\vx\in\mathfrak X} \Pr[M_\vx | \hat{m},\mathfrak D]$ next, which is the function input that is most likely %
 to achieve the highest function value.

Of course, the function maximum $m$ may be unknown. In this case, we use a plug-in estimate via 
the posterior expectation of $Y=\max_{\vx\in\mathfrak X} f(\vx)$ given $\mathfrak D$~\citep{ross2003useful}:
\begin{align}
\hat{m} &=  \mathbb  E[Y|\mathfrak D] \\ 
&= \int_0^{\infty} \Pr[Y>y|\mathfrak D] - \Pr[Y<-y|\mathfrak D]\dif y. \label{estm_exact}
\end{align}
If the noise in the observations is negligible, we can simplify Eqn.~\eqref{estm_exact} to be
\begin{align}
\hat{m}
&= m_0+\int_{m_0}^\infty 1- \prod_{\vx\in\mathfrak X} \Phi\Big(\frac{w-\mu(\vx)}{\sigma(\vx)}\Big) \dif w \label{estm}
\end{align}
where $m_0=\max_{\tau\in[1,t-1]}y_\tau$ is the current observed maximum value. Under conditions specified in Section~\ref{sec:bounds}, our approximation with the independence assumption makes $\hat m$ an upper bound on $m$, which, as we will see, conservatively emphasizes exploration a bit more. Other ways of setting $\hat m$ are discussed in Section~\ref{sec:disc}.

\hide{

Knowing $m$, the event $M_\vx|m,\mathfrak{D}$ is equivalent to %
\begin{align}
\Pr[M_\vx | m,\mathfrak D] %
& = Q\Big(\frac{m-\mu(\vx)}{\sigma(\vx)}\Big)\prod_{\vx'\neq \vx}\Phi\Big(\frac{m-\mu(\vx')}{\sigma(\vx')}\Big) \nonumber %
\end{align}
The algorithm GP-EST is described in Alg.~\ref{est}.
\
\begin{algorithm}[H]
  \caption{GP-EST$\max$}\label{est}
  \begin{algorithmic}[1]
  \State \textbf{Inputs:} sequence length $T$, blackbox function $f$, candidate set $\mathcal A$, sampled data $D$
  \State \textbf{Outputs:} sequence of sampled data points $x_t: $, $t = 1,...,T$
      \For{t = 1:$T$}
      \State $\vmu_t, \mSigma_t$ $\gets$ GP-predict($\mathcal A|D$)
      \State Estimate $Y$ to be $m$ with Eqn.~\eqref{estm}
      \State Compute $\Pr(M_i|m,D)$ with Eqn.~\eqref{pmi}
      \State Choose i that maximize $\Pr(M_i|m,D)$ 
      \State $\vx_{r(t)}$ $\gets$ $\vx_i$, test $y_{r(t)}=f(\vx_{r(t)})$
      \State $D\gets D\cup \{( \vx_{r(t)},y_{r(t)}  )\}$
      \EndFor
  \end{algorithmic}
\end{algorithm}
Our strategy (EST) plays~$\vx_t = \argmax_{\vx\in\mathfrak X} \Pr[M_\vx | \hat{m},\mathfrak D_t]$. 

}

\subsection{Connections to GP-UCB and GP-PI}\label{subsec:connections}
Next, we relate our strategy to GP-PI and GP-UCB: EST turns out to be equivalent to adaptively tuning $\theta_t$ in GP-PI and $\lambda_t$ in GP-UCB. This observation reveals unifying connections between GP-PI and GP-UCB and, in Section~\ref{sec:bounds}, yields regret bounds for GP-PI with a certain choice of $\theta_t$.
Lemma~\ref{thm1} characterizes the connection to GP-UCB:
\hide{
\begin{lem}
\label{thm1}
At any step $t$, the point selected by EST is the same as the point selected by a variant of GP-UCB with $\beta^{\frac12}_t = \min_{\vx\in\mathfrak X} \frac{\hat{m}-\mu_{t-1} (\vx)}{\sigma_{t-1} (\vx)}$.
\end{lem}
Conversely, GP-UCB may be understood as running EST with a specific choice of $\hat{m}$ (details in the supplement).}
\begin{restatable}{lem}{ucblemma}
\label{thm1}
In any round $t$, the point selected by EST is the same as the point selected by a variant of GP-UCB with $\lambda_t = \min_{\vx\in\mathfrak X} \frac{\hat{m}_t-\mu_{t-1} (\vx)}{\sigma_{t-1} (\vx)}$. Conversely, the candidate selected by GP-UCB is the same as the candidate selected by a variant of EST with $\hat m_t = \max_{\vx\in\mathfrak X}{\mu_{t-1} (\vx) + \lambda_t\sigma_{t-1}(\vx)}$.
\end{restatable}
\hide{
\begin{proof}
We omit the subscripts $t$ for simplicity. Let $\va$ be the point selected by GP-UCB, and $\vb$ the one selected by EST. Without loss of generality, we assume $\va$ and $\vb$ are unique. 
With $\beta^{\frac12} = \min_{\vx\in\mathfrak X} \frac{\hat{m}-\mu(\vx)}{\sigma(\vx)}$, GP-UCB plays $\va =\max_{\vx\in\mathfrak X}\mu(\vx) + \beta_t^{\frac12}\sigma(\vx)= \argmin_{\vx\in\mathfrak X} \frac{\hat{m}-\mu(\vx)}{\sigma(\vx)}$, because $$\hat m = \max_{\vx\in\mathfrak X} \mu(\vx) + \beta^{\frac12}\sigma(\vx)= \mu(\va) + \beta^{\frac12}\sigma(\va)$$
By definition of $\vb$, for all $\vx\in\mathfrak{X}$,  we have
\begin{align}
  \nonumber
1 \leq \frac{\Pr[M_{\vb} | \hat m,\mathfrak D]} {\Pr[M_\vx| \hat m,\mathfrak D]} &=
 \frac{Q(\frac{\hat{m}-\mu(\vb)}{\sigma(\vb)})\prod_{\vx'\neq \vb}\Phi(\frac{\hat{m}-\mu(\vx')}{\sigma(\vx')})}{Q(\frac{\hat{m}-\mu(\vx)}{\sigma(\vx)})\prod_{\vx'\neq \vx}\Phi(\frac{\hat{m}-\mu(\vx')}{\sigma(\vx')})} \\
 &=
  \frac{Q(\frac{\hat{m}-\mu(\vb)}{\sigma(\vb)}) \Phi(\frac{\hat{m}-\mu(\vx)}{\sigma(\vx)})}{Q(\frac{\hat{m}-\mu(\vx)}{\sigma(\vx)})\Phi(\frac{\hat{m}-\mu(\vb)}{\sigma(\vb)})} \label{prmneq}
\end{align}
The inequality holds if and only if $\frac{\hat{m}-\mu(\vb)}{\sigma(\vb)}\leq \frac{\hat{m}-\mu(\vx)}{\sigma(\vx)}$ for all $\vx\in\mathfrak X$, including $\va$, and hence
\begin{align}
\frac{\hat{m}-\mu(\vb)}{\sigma(\vb)}\leq \frac{\hat{m}-\mu(\va)}{\sigma(\va)} = \beta^{\frac12} = \min_{\vx\in\mathfrak X} \frac{\hat{m}-\mu (\vx)}{\sigma (\vx)} \label{thm2contra}
\end{align}
which, with uniqueness, implies that $\va=\vb$ and GP-UCB and EST select the same point.
\end{proof}
}

\begin{proof}
We omit the subscripts $t$ for simplicity. Let $\va$ be the point selected by GP-UCB, and $\vb$ selected by EST. Without loss of generality, we assume $\va$ and $\vb$ are unique. 
With $\lambda = \min_{\vx\in\mathfrak X} \frac{\hat{m}-\mu(\vx)}{\sigma(\vx)}$, GP-UCB chooses to evaluate 
$$\va =\argmax_{\vx\in\mathfrak X}\mu(\vx) + \lambda\sigma(\vx)= \argmin_{\vx\in\mathfrak X} \frac{\hat{m}-\mu(\vx)}{\sigma(\vx)}.$$
This is because $$\hat m = \max_{\vx\in\mathfrak X} \mu(\vx) + \lambda\sigma(\vx)= \mu(\va) + \lambda\sigma(\va).$$
By definition of $\vb$, for all $\vx\in\mathfrak{X}$,  we have
\begin{align}
\frac{\Pr[M_{\vb} | \hat m,\mathfrak D]} {\Pr[M_\vx| \hat m,\mathfrak D]} %
 &\approx
  \frac{Q(\frac{\hat{m}-\mu(\vb)}{\sigma(\vb)}) \Phi(\frac{\hat{m}-\mu(\vx)}{\sigma(\vx)})}{Q(\frac{\hat{m}-\mu(\vx)}{\sigma(\vx)})\Phi(\frac{\hat{m}-\mu(\vb)}{\sigma(\vb)})} \nonumber
  \geq 1 .
  \label{prmneq}
\end{align}
The inequality holds if and only if $\frac{\hat{m}-\mu(\vb)}{\sigma(\vb)}\leq \frac{\hat{m}-\mu(\vx)}{\sigma(\vx)}$ for all $\vx\in\mathfrak X$, including $\va$, and hence
\begin{align}
\frac{\hat{m}-\mu(\vb)}{\sigma(\vb)}\leq \frac{\hat{m}-\mu(\va)}{\sigma(\va)} = \lambda = \min_{\vx\in\mathfrak X} \frac{\hat{m}-\mu (\vx)}{\sigma (\vx)} ,\nonumber %
\end{align}
which, with uniqueness, implies that $\va=\vb$ and GP-UCB and EST select the same point.

The other direction of the proof is similar and can be found in the supplement.
\end{proof}

\begin{prop}\label{prop:pi-est}
GP-PI is equivalent to EST when setting $\theta_t = \hat m_t$ in GP-PI. 
\end{prop}
As a corollary of Lemma~\ref{thm1} and Proposition~\ref{prop:pi-est}, we obtain a correspondence between GP-PI and GP-UCB.
\begin{cor}\label{cor:ucb_pi}
 GP-UCB is equivalent to GP-PI if $\lambda_t$ is set to $\min_{\vx\in\mathfrak X} \frac{\theta_t-\mu_{t-1} (\vx)}{\sigma_{t-1} (\vx)}$, and GP-PI corresponds to GP-UCB if $\theta_t = \max_{\vx\in\mathfrak X}{\mu_{t-1} (\vx) + \lambda_t\sigma_{t-1}(\vx)}$.
\end{cor}

\hide{
\begin{algorithm}[H]
  \caption{GP-UCB/PI/EST %
  }\label{est}
  \begin{algorithmic}[1]
    \State $\mathfrak D \gets \emptyset$
      \While{\text{stopping criterion not reached}}
      \State $\mu, \Sigma$ $\gets$ GP-predict($\mathfrak X|\mathfrak D$)
      \State $\hat m$ = $\left\{
\begin{array}{l l}      
  \max_{\vx\in\mathfrak X}\mu(\vx) + \lambda\sigma(\vx) & \text{GP-UCB}\\
   \max_{1\leq\tau<t}y_\tau+\epsilon & \text{GP-PI}\\
 \mathbb  E[Y|\mathfrak D] \text{ (\text{see} Eqn.~\eqref{estm_exact}/Eqn.~\eqref{estm})} & \text{EST}
\end{array}\right.$
      \State $\vx_t\gets \argmin_{\vx\in\mathfrak X}{\frac{\hat{m}-\mu_{t-1} (\vx)}{\sigma_{t-1} (\vx)}}$ 
      \State $y\gets f(\vx)+ \epsilon$%
      \State $\mathfrak D \gets D\cup\{(\vx,y)\}$
      \EndWhile
  \end{algorithmic}
\end{algorithm}
}

\begin{algorithm}[H]
  \caption{GP-UCB/PI/EST %
  }\label{est}
  \begin{algorithmic}[1]
  \small
    \State $t\gets 1; \mathfrak D_0 \gets \emptyset$
      \While{\text{stopping criterion not reached}}
      \State $\mu_{t-1}, \Sigma_{t-1}$ $\gets$ GP-predict($\mathfrak X|\mathfrak D_{t-1}$)
      \State $\hat m_t$ = $\left\{
\begin{array}{l l}      
  \max_{\vx\in\mathfrak X}\mu_{t-1}(\vx) + \lambda_t\sigma_{t-1}(\vx) & \text{GP-UCB}\\
   \max_{1\leq\tau<t}y_\tau+\epsilon & \text{GP-PI}\\
 \mathbb  E[Y|\mathfrak D_{t-1}] \text{ (\text{see} Eqn.~\eqref{estm_exact}/Eqn.~\eqref{estm})} & \text{EST}
\end{array}\right.$
      \State $\vx_t\gets \argmin_{\vx\in\mathfrak X}{\frac{\hat{m}_t-\mu_{t-1} (\vx)}{\sigma_{t-1} (\vx)}}$ 
      \State $y_t\gets f(\vx_t) + \epsilon_t, \epsilon_t\sim\mathcal N(0,\sigma^2)$
      \State $\mathfrak D_t \gets \{\vx_\tau,y_\tau\}_{\tau=1}^{t}$
      \State $t\gets t+1$
      \EndWhile
  \end{algorithmic}
\end{algorithm}

Proposition~\ref{prop:pi-est} suggests that we do not need to calculate the probability $\Pr[M_\vx | \hat m_t,\mathfrak D_{t-1}]$ directly when implementing EST. Instead, we can reduce EST to GP-PI with an automatically tuned target value $\theta_t$.  Algorithm~\ref{est} compares the pseudocode for all three methods. We use ``GP-predict'' to denote the update for the posterior mean and covariance function for the GP as described in Section~\ref{sec:background}. 
 GP-UCB/PI/EST all share the same idea of reaching a target value ($\hat m_t$ in this case), and thereby trading off exploration and exploitation. GP-UCB in~\cite{srinivas2009gaussian} can be interpreted as  setting the target value to be a loose upper bound $ \max_{\vx\in\mathfrak X}\mu_{t-1}(\vx) + \lambda_t\sigma_{t-1}(\vx)$ with $\lambda_t= (2\log(|\mathfrak X|\pi^2 t^2/6\delta))^\frac12$
, as a result of applying the union bound over $\mathfrak X$\footnote{Since $\Pr[|f(\vx) - \mu(\vx)|>\lambda_t\sigma(\vx)]\leq e^{\frac{-\lambda_t^2}{2}}$, applying the union bound results in $\Pr[|f(\vx) - \mu(\vx)|>\lambda_t\sigma(\vx), \forall \vx\in\mathfrak X]\leq |\mathfrak X|e^{\frac{-\lambda_t^2}{2}}$. This means $f(\vx)\leq \max_{\vx\in\mathfrak X}\mu(\vx) + \lambda_t\sigma(\vx)$ with probability at least $1-|\mathfrak X|e^{\frac{-\lambda_t^2}{2}}$~\citep[Lemma 5.1]{srinivas2009gaussian}.}. %
GP-PI applies a fixed upwards shift of $\epsilon$ over the current maximum observation $\max_{\tau\in[1,t-1]}y_\tau$. In both cases, the exploration-exploitation tradeoff depends on the parameter to be set.
EST implicitly and automatically balances the two by estimating the maximum. Viewed as GP-UCB or GP-PI, it automatically sets the respective parameter.

Note that this change by EST is not only intuitively reasonable, but it also leads to vanishing regret, %
as will become evident in the next section. 

\section{Regret Bounds}\label{sec:bounds}

In this section, we analyze the regret of EST.  %
We first show a bound on the cumulative regret both in expectation and with high probability, with the assumption that our estimation $\hat m_t$ is always an upper bound on the maximum of the function. Then we interpret how this assumption is satisfied via Eqn.~\eqref{estm_exact} and Eqn.~\eqref{estm} under the condition specified in Corollary \ref{hatm_upperbound}.

\hide{
\begin{thm}\label{thm:regretbound}
We assume $\hat m_t\geq\max_{\vx\in \mathfrak X} f(\vx), \forall t\in[1,T]$, and restrict $k(\vx,\vx')\leq 1$. Let $\sigma^2$ be the variance of the Gaussian noise in the observation,
 $C = 2/\log (1+\sigma^{-2})$,  $\gamma_T$ the maximum information gain of the selected points, and $t^*=\argmax_t \bt_t$ where $\bt_t^{\frac12} \triangleq \min_{\vx\in\mathfrak X} \frac{\hat m-\mu_{t-1}(\vx)}{\sigma_{t-1}(\vx)}$.
With probability $1-\delta$, the cumulative regret up to time step $T$ is bounded as
\begin{equation*}
  R_T = \frac1T\sum_{t=1}^T \rt_t \leq \sqrt{\frac{C \gamma_T}{T}} (\bt^{\frac12}_{t^*} + \zeta^{\frac12}_T), 
\end{equation*}
where $\zeta_T$ is defined in Lemma~\ref{lem:pbound}.
\end{thm}}
\begin{thm}\label{thm:regretbound}
We assume $\hat m_t\geq\max_{\vx\in \mathfrak X} f(\vx), \forall t\in[1,T]$, and restrict $k(\vx,\vx')\leq 1$. 
Let $\sigma^2$ be the variance of the Gaussian noise in the observation,  $\gamma_T$ the maximum information gain of the selected points%
, $C = 2/\log (1+\sigma^{-2})$, and $t^*=\argmax_t \bt_t$ where $\bt_t \triangleq \min_{\vx\in\mathfrak X} \frac{\hat m_t-\mu_{t-1}(\vx)}{\sigma_{t-1}(\vx)}$. %
The cumulative expected  regret satisfies
$
\sum_{t=1}^T  \mathbb E\left[\rt_t|\mathfrak D_{t-1}\right] \leq \bt_{t^*} \sqrt{C T\gamma_T  }.
$
With probability at least $1-\delta$, it holds that %
$
   \sum_{t=1}^T \rt_t\leq (\bt_{t^*} + \zeta_T)\sqrt{C T\gamma_T} , 
$
with $\zeta_{T} = (2\log(\frac{T}{2\delta}))^\frac12$.%
\end{thm}
The information gain $\gamma_T$ after $T$ rounds is the maximum mutual information that can be gained about $f$ from $T$ measurements: $\gamma_T = \max_{A \subseteq \mathfrak{X}, |A| \leq T} I(\vy_A, \vf_A) = \max_{A \subseteq \mathfrak{X}, |A| \leq T} \frac{1}{2}\log \det(\mI + \sigma^{-2}\mK_A)$. %
For the Gaussian kernel, $\gamma_T = O((\log T)^{d+1})$, and for the Mat\'ern kernel, $\gamma_T = O(T^{d(d+1)/(2\xi+d(d+1))}\log T)$ where $d$ is the dimension and $\xi$ is the roughness parameter of  the kernel~\cite[Theorem 5]{srinivas2009gaussian}. %

The proof of Theorem~\ref{thm:regretbound} follows ~\cite{srinivas2009gaussian} and  relies on the following lemmas which are proved in the supplement. %

\begin{restatable}{lem}{pbound}\label{lem:pbound}
Pick $\delta\in(0,1)$ and set $\zeta_{t} = (2\log(\frac{\pi_t}{2\delta}))^\frac12$, where $\sum_{t=1}^T \pi_t^{-1} \leq 1$, $\pi_t > 0$. Then, for EST, it holds that
$ \Pr [  \mu_{t-1}(\vx_{t}) -f( \vx_{t})  \leq \zeta_{t}\sigma_{t-1}(\vx_{t}) ] \geq 1-\delta$, for all $t\in [1,T]$.
\end{restatable}
Lemma \ref{lem:pbound} is similar to but not exactly the same as~\citep[Appendix A.1]{srinivas2009gaussian}:
while they use a union bound over all of $\mathfrak X$, here, we only need a union bound over the actually evaluated points $\{\vx_t\}_{t=1}^T$. This difference is due to the different selection strategies.
\begin{restatable}{lem}{regretlem}
\label{regretlem}
If $ \mu_{t-1}(\vx_{t}) -f(\vx_{t}) \leq \zeta_t\sigma_{t-1}(\vx_{t})$, the regret at time step $t$ is upper bounded as $\rt_t \leq (\bt_t  +\zeta_t )\sigma_{t-1}(\vx_{t})$ %
, where $\bt_t  \triangleq \min_{\vx\in\mathfrak X} \frac{\hat m_t-\mu_{t-1}(\vx)}{\sigma_{t-1}(\vx)}$, and $\hat m_t\geq \max_{\vx\in \mathfrak X} f(\vx)$, $\forall t\in[1,T]$.
\end{restatable}

The proof of Theorem~\ref{thm:regretbound} now follows from Lemmas~\ref{lem:pbound}, \ref{regretlem}, and Lemma 5.3 in~\citep{srinivas2009gaussian}.

\begin{proof}\textit{(Thm.~\ref{thm:regretbound})}
\hide{
By Lemma~\ref{regretlem}, we have that $\rt_t^2 \leq (\bt_{t^*}^{\frac12} +\zeta_{T}^{\frac12})^2 \sigma^2_{t-1}(\vx_{t})$. Using $(1+a)^x\leq 1+ax$ if $0\leq x\leq1$, it follows that
\begin{equation*}
 \sigma^2_{t-1}(\vx_{t}) \leq \frac{\log (1+\sigma^{-2}\sigma^2_{t-1}(\vx_{t}))}{\log (1+\sigma^{-2})}.
\end{equation*}
By Lemma 5.3 and Lemma 5.4 in~\cite{srinivas2009gaussian}, we further bound
$\sum_{t=1}^T  \sigma^2_{t-1}(\vx_{t}) \leq \frac{2}{\log(1+\sigma^{-2})} \gamma_T$.
Now, by the Cauchy-Schwarz inequality, we obtain
\begin{equation*}
  \sum_{t=1}^T\sigma_{t-1}(\vx_{t}) \leq \sqrt{T\sum_{t=1}^T  \sigma^2_{t-1}(\vx_{t}) }\leq \sqrt{\frac{2T}{\log(1+\sigma^{-2})} \gamma_T}.
\end{equation*}
and therefore
\begin{align*}
  \sum_{t=1}^T \rt_t &\leq  \sum_{t=1}^T\sigma_{t-1}(\vx_t)(\bt_{t^*}^{\frac12} + \zeta_T^{\frac12}) \\ &\leq \sqrt{\frac{2T \gamma_T}{\log (1+\sigma^{-2})}} (\bt_{t^*}^{\frac12} + \zeta_T^{\frac12}) \qedhere
\end{align*}
}
The expected regret of %
round $t$ is
$
\Ex[\rt_t|\mathfrak D_{t-1}] %
\leq \hat m_t - \mu_{t-1}(\vx_{t}) 
= \bt_t  \sigma_{t-1}(\vx_{t}).
$
Using $t^* = \argmax_t \nu_t$, we obtain that
$
\sum_{t=1}^T\mathbb E\left[\rt_t\right|\mathfrak D_{t-1}] \leq \nu_{t^*}\sum_{t=1}^T \sigma_{t-1}(\vx_t).
$

To bound the sum of variances, we first use that $(1+a)^x\leq 1+ax$ for $0\leq x\leq1$ and the assumption $\sigma_{t-1}(\vx_t)\leq k(\vx_t,\vx_t)\leq 1$ to obtain
$
 \sigma^2_{t-1}(\vx_{t}) \leq \frac{\log (1+\sigma^{-2}\sigma^2_{t-1}(\vx_{t}))}{\log (1+\sigma^{-2})}.
$
Lemma 5.3 in~\citep{srinivas2009gaussian} now implies that $\sum_{t=1}^T  \sigma^2_{t-1}(\vx_{t}) \leq \frac{2}{\log(1+\sigma^{-2})} I(\vy_T;\vf_T )\leq  \frac{2}{\log(1+\sigma^{-2})} \gamma_T$. %
The Cauchy-Schwarz inequality leads to
$
  \sum_{t=1}^T\sigma_{t-1}(\vx_{t}) \leq \sqrt{T\sum_{t=1}^T  \sigma^2_{t-1}(\vx_{t}) }\leq \sqrt{\frac{2T\gamma_T}{\log(1+\sigma^{-2})} }.
$
Together, we have the final regret bound 
$$
 \sum_{t=1}^T\mathbb E\left[\rt_t\right|\mathfrak D_{t-1}] 
\leq\bt_{t^*}\sqrt{\frac{2T }{\log(1+\sigma^{-2})}\gamma_T}.
$$

Next we show a high probability bound. The condition of Lemma~\ref{regretlem} holds with high probability because of Lemma~\ref{lem:pbound}. Thus with probability at least $1-\delta$, the regret for round $t$ is bounded as follows,%
$$\rt_t \leq (\bt_t  +\zeta_t )\sigma_{t-1}(\vx_{t})\leq (\bt_{t^*} +\zeta_{T} )\sigma_{t-1}(\vx_{t}),$$ 
where $\zeta_{t} = 2\log(\frac{\pi_t}{2\delta})$, $\pi_t = \frac{\pi^2 t^2}{6}$, and $t^*=\argmax_t \bt_t$. Therefore, with probability at least $1-\delta$,
\begin{align*}
  \sum_{t=1}^T \rt_t %
  \leq (\bt_{t^*} + \zeta_T)\sqrt{\frac{2T \gamma_T}{\log (1+\sigma^{-2})}}.  \;\;\;\;\;\;\;
  \qedhere
\end{align*}

\end{proof}

\hide{
\begin{lem}
\label{expdlem}
For EST, the expected regret at time step $t$ is $\mathbb{E}[\rt_t] = \bt_t^{\frac12} \sigma_{t-1}(\vx_{t})$. The expected simple regret is upper bounded by $\mathbb{E}[r_t] = \mathbb{E}[\min_{t\in[1,T]} \rt_t] \leq \bt_T^{\frac12} \sigma_{T-1}(\vx_{T})$.
\end{lem}
}
\hide{
Theorem~\ref{thm:regretbound} directly implies a convergence bound on the number of rounds needed to achieve a regret of at most $\epsilon$.
\begin{cor}[Trial Bound]\label{cor:trialbound}
In expectation, within $T \geq C  \bt_{t^*} \gamma_T / \epsilon^2$ rounds, EST achieves, a simple regret or average cumulative regret of at most $\epsilon$;
within $T \geq C  (\bt_{t^*} + \zeta_T + 2(\bt_{t^*}\zeta_T)^{\frac12}) \gamma_T / \epsilon^2$ rounds, EST achieves, with probability $1-\delta$, a simple regret or average cumulative regret of at most $\epsilon$, where $C, \gamma_T, \bt_{t^*}, \zeta_T$ are defined in Theorem~\ref{thm:regretbound}.
\end{cor}
}
Next we show that if we estimate $\hat m_t$ as described in Section~\ref{sec:method} by assuming all the $\{f(\vx)\}_{\vx\in\mathfrak X}$ are independent conditioned on the current sampled data $\mathfrak{D}_t$, $\hat m_t$ can be guaranteed to be an upper bound on the function maximum $m$ given $k_t(\vx,\vx')\geq 0, \forall \vx,\vx'\in \mathfrak X$.

\begin{lem}[Slepian's Comparison Lemma~\cite{slepian1962one,massart2007concentration}]
Let $\vu, \vv\in \R^n$ be two multivariate Gaussian random vectors with the same mean and variance, such that
 $\mathbb E [\vv_i\vv_j]\leq \mathbb E [\vu_i\vu_j], \forall i,j.$
 Then
 $\mathbb E [\sup_{i\in[1,n]}\vv_i] \geq \mathbb E [\sup_{i\in[1,n]}\vu_i].$
\end{lem}

Slepian's Lemma implies a relation between our approximation $\hat{m}_t$ and $m$.%

\begin{cor}\label{hatm_upperbound}
Assume $g\sim GP(\mu,k)$ has posterior mean $\mu_t(\vx)$ and posterior covariance $k_{t}(\vx,\vx')\geq 0, \forall \vx,\vx'\in \mathfrak X$ conditioned on $\mathfrak{D}_t$. Define a series of \emph{independent} random variables $h(\vx)$ with equivalent mean $\mu_t(\vx)$ and posterior variance $k_t(\vx,\vx)$, $\forall \vx\in\mathfrak X$. Then, $\mathbb E [\sup_{\vx\in \mathfrak X} h(\vx)] \geq \mathbb  E [\sup_{\vx\in \mathfrak X} g(\vx)]$.
\end{cor}
\begin{proof}
By independence, $\forall \vx,\vx'\in \mathfrak X$ 
\begin{align*}
0 &=\mathbb E[h(\vx)h(\vx')] -\mathbb E[h(\vx)]\mathbb E[h(\vx')]\\
&\leq E[g(\vx)g(\vx')]-\mathbb E[g(\vx)]\mathbb E[g(\vx')]
\end{align*}
Hence, Slepian's Lemma implies that $\mathbb E [\sup_{\vx\in \mathfrak X} h(\vx)] \geq \mathbb  E [\sup_{\vx\in \mathfrak X} g(\vx)]$.
\end{proof}
Corollary~\ref{hatm_upperbound} assumes that $k_{t}(\vx,\vx')\geq 0$, $\forall \vx,\vx'\in \mathfrak X$. This depends on the choice of $\mathfrak X$ and $k$. %
Notice that $k_{t}(\vx,\vx')\geq 0$ is only a sufficient condition and, even if the assumption fails, $\hat m_t$ is often still an upper bound on $m$ in practice (illustrations in the supplement). 

In contrast, the results above are not necessarily true for any arbitrary $\theta_t$ in GP-PI, an important distinction between GP-PI and GP-EST.

Before evaluating the EST strategy empirically, %
we make a few important observations.
First, EST does not require manually setting a parameter that trades off exploration and exploitation. Instead, it corresponds to automatically, adaptively setting the tradeoff parameters $\lambda_t$ in GP-UCB and $\theta_t$ in GP-PI. %
For EST, this means that if the gap $\bt_t$ is large, then the method focuses more on exploration, and if ``good'' function values (i.e., close to $\hat m_t$) are observed, then exploitation increases. %
 If we write EST as GP-PI, we see that by Eqn.~\eqref{estm}, the estimated $\hat{m}_t$ always ensures $\theta_t > \max_{\tau\in[1,T]} y_\tau$, which is known to be advantageous in practice~\cite{lizotte2012}. These analogies likewise suggest that $\theta_t = \max_{\tau\in[1,T]} y_\tau$ corresponds to a very small $\lambda_t$ in GP-UCB, and results in very little exploration, offering an explanation for the known shortcomings of this $\theta_t$.

\section{Experiments}
\label{sec:exp}

We test EST\footnote{The code is available at \scriptsize{\url{https://github.com/zi-w/GP-EST}}.} in three domains: (1) synthetic black box functions; (2) initialization tuning for trajectory optimization; and (3) parameter tuning for image classification.
We compare the following methods: 
EST with a Laplace approximation, i.e., approximating the integrand in Eqn.~\eqref{estm} by a truncated Gaussian (\ESTa, details in supplement); EST with numerical integration to evaluate Eqn.~\eqref{estm} (\ESTn); \UCB; \EI; \PI; and random selection (\RND). We omit the 'GP-' prefix for simplicity.

For \UCB, we follow~\cite{srinivas2009gaussian} to set $\lambda_t$ with $\delta = 0.01$. For \PI, we use $\epsilon = 0.1$. These parameters are  coarsely tuned via cross validation to ensure a small number of iterations for achieving low regret. Additional experimental results and details may be found in the supplement.

\subsection{Synthetic Data}
\label{sec:synthexp}
\begin{figure*}
\centering
\includegraphics[width=0.94\textwidth]{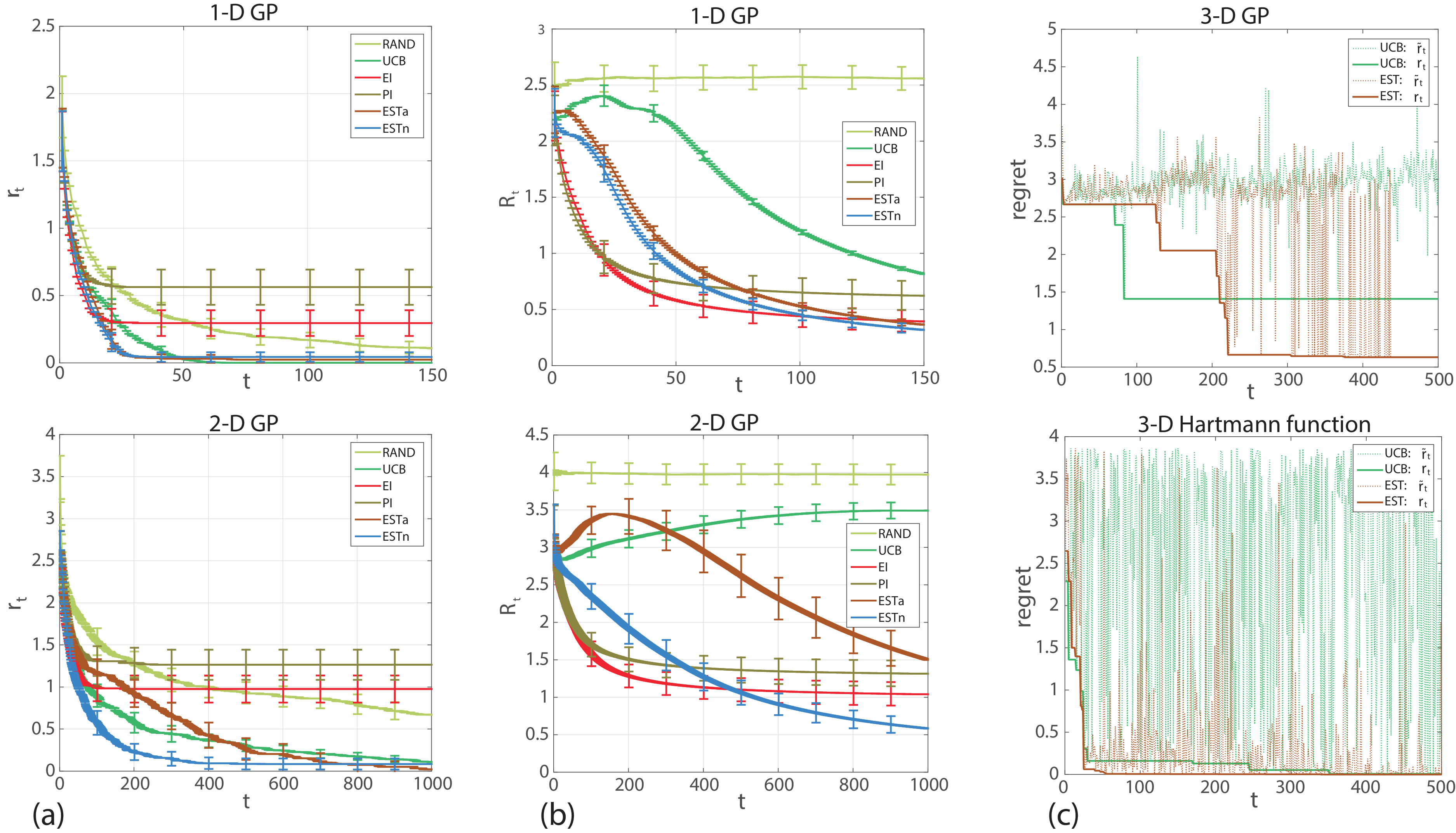}
\caption{(a) Simple regrets for functions sampled from 1-D GP and 2-D GP over the number of rounds. \ESTn\ quickly minimizes regret. (b) Cumulative regrets for functions sampled from 1-D GP and 2-D GP. \ESTn\ lowers the regret more than \PI\ and \EI\, and faster than \UCB. (c) Regrets $\tilde r_t$ and simple regrets $r_t$ for a 3-D function sampled from GP and the 3-D Hartmann function. Here, \EST\ is \ESTa. For \EST, the regret in each round is usually less than that of \UCB, explaining \UCB's higher cumulative regret.}
\label{fig:synth}
\end{figure*}

We sampled 200 functions from a 1-D GP and 100 functions from a 2-D GP with known priors (Mat\'ern kernel and linear mean function). The maximum number of rounds was 150 for 1-D and 1000 for 2-D. The first samples were the same for all the methods to minimize randomness effects. Table~\ref{tab:1dgp} shows the lowest simple regret achieved ($r_{\min}$) and the number of rounds needed to reach it ($T_{\min}$). We measure the mean ($\bar T_{\min}$ and $\bar r_{\min}$) and the median ($\hat T_{\min}$ and $\hat r_{\min}$).
Figure~\ref{fig:synth}(a) illustrates the average simple regret and the standard deviation (scaled by $1/4$).
While the progress of \EI\ and \PI\ quickly levels off, the other methods continue to reduce the regret, \ESTn\ being the fastest. Moreover, the standard deviation of \ESTa\ and \ESTn\ is much lower than that of \PI\ and \EI.
\RND\ is inferior both in terms of $T_{\min}$ and $r_{\min}$. \UCB\ finds a good point but takes more than twice as many rounds as \EST\ for doing so.

\begin{table}\caption{Minimum Regret $r(t)$ and time to achieve this minimum for functions sampled from 1-D (top) and 2-D Gaussian Processes (bottom) with a limited budget of iterations. 
  \ESTa\ and \ESTn\ achieve lower regret values ($r_{\min}$) faster than other methods ($T_{\min}$). %
Here, `\={}' denotes the mean and `\^{}' the median. %
}
    \label{tab:1dgp}
{ \small
\begin{center}
    \begin{tabular}{ |c|c|c|c|c|c|c|}
    \hline
    \multicolumn{7}{|c|}{1-D GP, max 150 rounds}\\\hline
     & RAND & UCB & EI & PI & ESTa & ESTn \\\hline
$\hat T_{min}$ & 79.5 & 53 & 8 & 7 & 26 & 23 \\
$\hat r_{min}$ & 0.051 & 0.000 & 0.088& 0.487 & 0.000 & 0.000 \\ 
$\bar T_{min}$ & 78.4 & 50.9 & 9.77 & 8.32 & 26.1 & 21.9 \\
$\bar r_{min}$ & 0.107 & 0.000 & 0.295 & 0.562 & 0.024 & 0.043 \\ 
\hline\hline
\multicolumn{7}{|c|}{2-D GP, max 1000 rounds}\\
    \hline
     & RAND & UCB & EI & PI & ESTa & ESTn \\\hline
$\hat T_{\min}$ & 450.5 & 641.5 & 40.5 & 45 & 407.5 & 181 \\
$\hat r_{\min}$ & 0.640 & 0.090 & 1.035 & 1.290 & 0.000 & 0.000 \\ 
$\bar T_{\min}$ & 496.3 & 573.8 & 48.4 & 59.8 & 420.7 & 213.4 \\
$\bar r_{\min}$ & 0.671 & 0.108 & 0.976 & 1.26 & 0.021 & 0.085 \\ \hline
    \end{tabular}
    \end{center}
}

\end{table}

Figure~\ref{fig:synth}(b) shows the cumulative regret $R_T$. As for the simple regret, we see that \EI\ and \PI\ focus too much on exploitation, stalling at a suboptimal point. \EST\ converges to lower values, and faster than \UCB.
For additional intuition on cumulative regret, Figure~\ref{fig:synth}(c) plots the cumulative regret $\rt_t$ and the simple regret $r_t$ for \UCB\ and \EST\ for a function sampled from 3-D GP and a standard optimization test function (the 3-D Hartmann function). UCB tends to have higher cumulative regret than other methods because it keeps exploring drastically even after having reached the optimum of the function. This observation is in agreement with the experimental results in~\cite{srinivas2009gaussian} (they improved UCB's performance by scaling $\lambda_t$ down by a factor of 5), indicating that the scheme of setting $\lambda_t$ in UCB is not always ideal in practice. 

In summary, the results for synthetic functions suggest that throughout, compared to other methods, \EST\ finds better function values within a smaller number of iterations.
\hide{
\begin{figure}
  \centering
  \includegraphics[width=0.4\textwidth]{figs/matern1d}
  \hspace{20pt}
  \includegraphics[width=0.4\textwidth]{figs/matern2d}

        \caption{Simple regrets for functions sampled from 1-D GP (left) and 2-D GP (right) over the number of rounds. \EST\ quickly minimizes regret.}
        \label{fig:gpres}
\end{figure}

\begin{figure}
        \centering
        \includegraphics[width=0.4\textwidth]{figs/matern1daccu}
        \hspace{20pt}
        \includegraphics[width=0.4\textwidth]{figs/matern2daccu}

        \caption{Cumulative regret for functions sampled from 1-D GP (left) and 2-D GP (right). \ESTn\ lowers the regret more than \PI\ and \EI\, and faster than \UCB.        \label{fig:gparc}}
\end{figure}

\begin{figure*}
        \centering
        \includegraphics[width=0.4\textwidth]{figs/3dgp}
        \hspace{20pt}
        \includegraphics[width=0.4\textwidth]{figs/hart3}

        \caption{Regrets $\tilde r_t$ and simple regrets $r_t$ for a 3-D function sampled from GP (left) and the 3-D Hartmann function. Here, \EST\ is \ESTa. For \EST, the regret in each round is usually less than that of \UCB, explaining \UCB's higher accumulated regret.}
        \label{fig:gp3d}
\end{figure*}
}

\subsection{Initialization Tuning for Trajectory Optimization}
In online planning, robots must make a decision quickly, within a possibly unknown budget (humans can stop the ``thinking period'' of the robot any time, asking for a feasible and good decision to execute). We consider the problem of trajectory optimization, a
non-convex optimization problem that is commonly solved via sequential quadratic programming (SQP) 
\cite{schulman2013finding}. %
The employed solvers suffer from sub-optimal local optima, 
and, in real-world scenarios, even merely reaching a feasible solution can be challenging. Hence, we use Bayesian Optimization to tune the initialization for trajectory optimization.
In this setting, $\vx$ is a trajectory initialization, and $f(\vx)$ the score of the solution returned by the solver after starting it at $\vx$.

\begin{table}\caption{Lowest reward attained in 20 rounds on the airplane problem (illustrated in Figure~\ref{fig:apreward}). The results are averages over 8 settings.}
    \label{tb:airplane}
\centering
    \begin{tabular}{ |c|c|c|c|}
    \hline
     & RAND & UCB & EI  \\\hline\hline
mean & 27.4429 & 29.2021 & 29.1832  \\
std  & 3.3962 & 3.1247 & 3.1377 \\ \hline
\hline
     &  PI & ESTa & ESTn \\\hline\hline
mean & 28.0214 & 27.7587 &29.2071 \\
std  & 4.1489 & 4.2783 & 3.1171 \\ \hline
\end{tabular}
\vspace{-5pt}
\end{table}

\begin{figure*}
\centering
\includegraphics[width=0.65\textwidth]{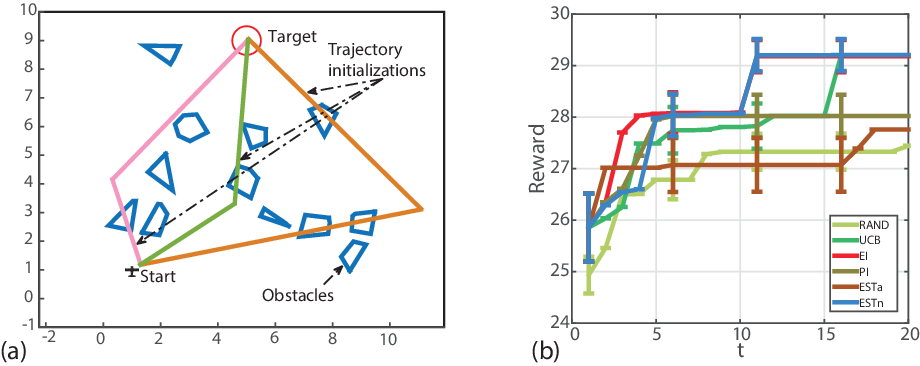}
\caption{(a) Illustration of trajectory initializations. Trajectory initializations are passed to the non-convex optimization solver to ensure a solution with as few collisions with the obstacles as possible.
(b) Maximum rewards up to round $t$. EI and ESTn perform relatively better than other methods.}
\label{fig:apreward}
\end{figure*}

Our test example is the 2D airplane problem from~\cite{drake}, illustrated in Figure~\ref{fig:apreward}. 
We used 8 configurations of the starting point and fixed the target. Our candidate set $\mathfrak X$ of initializations is a grid of the first two dimensions of the midpoint state of the trajectory (we do not optimize over speed here). To solve the SQP, we used SNOPT~\cite{gill2002snopt}.
Figure~\ref{fig:apreward} shows the maximum rewards achieved up to round $t$ (standard deviations scaled by 0.1), and Table~\ref{tb:airplane} displays the final rewards. \ESTn\ achieves rewards on par with the best competitors. Importantly, we observe that Bayesian optimization achieves much better results than the standard random restarts, indicating a new successful application of Bayesian optimization.
\hide{
\begin{figure}
  \centering
  \begin{minipage}[m]{0.46\linewidth}
    \includegraphics[height=0.75\textwidth]{figs/plane}
  \end{minipage}
  \begin{minipage}[m]{0.46\linewidth}
    \includegraphics[height=0.7\textwidth]{figs/planeres}
  \end{minipage}
        \vspace{-5pt}
        \caption{Left: Illustration of trajectory initializations with (fixed) starting position $S$ and target $T$. Right: Maximum rewards up to round $t$. Both EI and ESTn}
\end{figure}
}

\vspace{-5pt}

\subsection{Parameter Tuning for Image Classification}
Our third set of experiments addresses Bayesian optimization for efficiently tuning parameters in visual classification tasks. Here, $\vx$ is the %
model parameter and $y$ the accuracy on the validation set. 
Our six image datasets are standard benchmarks for object classification (Caltech101~\cite{fei2007learning} and Caltech256~\cite{griffin2007caltech}), scene classification (Indoor67~\cite{quattoni2009recognizing} and SUN397~\cite{xiao2010sun}), and action/event classification (Action40~\cite{yao2011human} and Event8~\cite{li2007and}). 
The number of images per data set varies from 1,500 to 100,000. We use deep CNN features pre-trained on ImageNet~\cite{Jia13caffe}, the state of the art on various visual classification tasks 
\cite{razavian2014cnn}. 

Our experimental setup follows~\cite{zhou2014learning}. %
The data is split into training, validation (20\% of the original training set) and test set following the standard settings of the datasets. %
We train a linear SVM using the deep features, and tune its regularization parameter $C$ via Bayesian optimization on the validation set. 
After obtaining the parameter recommended by each method, we train the classifier on the whole training set, and then evaluate on the test set.

\begin{figure*}
\begin{center}
\includegraphics[width=0.94\linewidth]{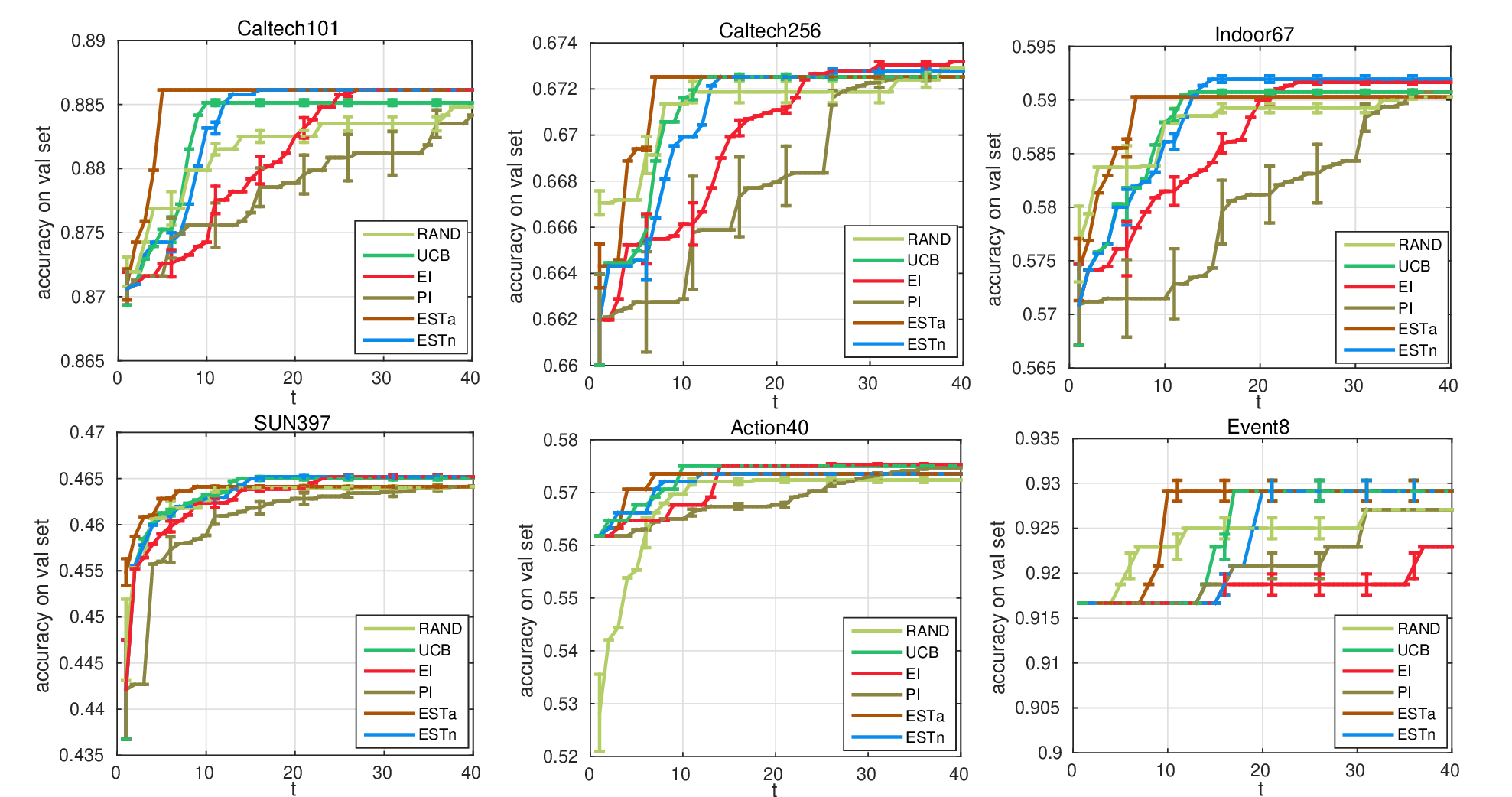}
\end{center}
   \caption{Maximum accuracy on the validation set over the iteration of the optimization. ESTa converges faster than other methods. Experiments are repeated 5 times, the standard deviation is scaled by 1/4.}\label{onvalidationset}
\end{figure*}

Figure \ref{onvalidationset} shows the maximum achieved accuracy on the validation set during the iterations of Bayesian optimization on the six datasets. While all methods improve the classification accuracy, \ESTa\ does so faster than other methods. Here too, \PI\ and \EI\ seem to explore too little.
Table \ref{ontestset} displays the accuracy on the test set using the best parameter found by \ESTa\ and \ESTn, indicating that the parameter tuning via \EST\ improved classification accuracy. For example, the tuning improves the accuracy on Action40 and SUN397 by 3-4\% over the results in~\cite{zhou2014learning}.

\begin{table}\caption{Classification accuracy for visual classification on the test set after the model parameter is tuned. Tuning achieves good improvements over the results in~\cite{zhou2014learning}.}\label{hybridnet}
\centering
\begin{tabular}{|c|c|c|c|}
\hline
& Caltech101 & Caltech256 & Indoor67 \\
\hline
\hline
\cite{zhou2014learning} & 87.22 & 67.23 & 56.79  \\
\hline
ESTa & 88.23 & 69.39 & 60.02 \\
ESTn & 88.25 & 69.39 & 60.08 \\
\hline
\hline
& SUN397 & Action40 & Event8 \\
\hline
\hline
\cite{zhou2014learning} & 42.61 & 54.92 & 94.42 \\
\hline
ESTa &  47.13 & 57.60 & 94.91\\
ESTn &  47.21 & 57.58 & 94.86\\
\hline
\end{tabular}
\vspace{-5pt}
\label{ontestset}
\end{table}

\vspace{-5pt}

\section{Discussion}
\label{sec:disc}
\vspace{-10pt}

Next, we discuss 
a few further details and extensions.
\paragraph{Setting $\hat m$.}
In Section~\ref{sec:method}, we discussed one way of setting $\hat m$. There, we used an approximation with independent variables. We focused on Equations~\eqref{estm_exact} and \eqref{estm} throughout the paper since they yield upper bounds $\hat{m} \geq m$ that preserve our theoretical results. Nevertheless, other possibilities for setting $\hat{m}$ are conceivable.
For example, close in spirit to Thompson and importance sampling, one may sample $\hat{m}$ from $\Pr[Y] = \prod_{\vx\in\mathfrak X}\Phi(\frac{Y-\mu(\vx)}{\sigma(\vx)})$. Furthermore, other search strategies can be used, such as guessing or doubling, and prior knowledge about properties of $f$ such as the range can be taken into account. %

\paragraph{Discretization. } 
In our approximations, we used discretizations of the input space. While adding a bit more detail about this step here, we
focus on the noiseless case, described by Equation~\eqref{estm}. Equation~\eqref{estm_exact} can be analyzed similarly for more general settings. 
For a Lipschitz continuous function, it is essentially sufficient to estimate
the probability of $f(\vx)\leq y, \forall \vx\in\mathfrak X$ %
on set $\mathfrak W$, which is a $\rho$-covering of $\mathfrak X$.

We assume that $f$ is Lipschitz continuous. Our analysis can be adapted to the milder assumption that $f$ is Lipschitz continuous with high probability. 
Let $L$ be the Lipschitz constant of $f$. By assumption, we have 
$|f(\vx) - f(\vx')|\leq L\rho$, for all $\|\vx-\vx'\|\leq \rho.$
If $\mathfrak{X}$ is a continuous set, we construct its $\rho$-covering $\mathfrak W$ such that $\forall \vx \in \mathfrak X$, $\inf_{\vx'\in \mathfrak W}\|\vx-\vx'\| \leq \rho$. %
Let $E_{\mathfrak X}(y)$ be the event that $f(\vx)\leq y, \forall \vx\in\mathfrak X$. %
 Then,
$
\Pr[E_{\mathfrak X}(y)] \geq \Pr[E_{\mathfrak W}(y-\rho L),E_{\mathfrak X\setminus \mathfrak W}(y) ] 
= \Pr[E_{\mathfrak W}(y-\rho L)] %
$
\hide{
 Then,
\begin{align*}
\Pr[E_{\mathfrak W}(y)] & \geq \Pr[E_{\mathfrak W'}(y-\epsilon L),E_{\mathfrak W\setminus \mathfrak W'}(y) ] \\
&\geq \Pr[E_{\mathfrak W'}(y-\epsilon L)] (1-|\mathfrak W'|dae^{-L^2/b^2})
\end{align*}
}
The last step uses Lipschitz continuity  to compute $\Pr[E_{\mathfrak X\setminus \mathfrak W}(y)|[E_{\mathfrak W}(y-\rho L)] = 1$. 
We can use this lower bound to compute $\hat m$, so $\hat m$ remains an upper bound on $m$. %
Notably, none of the regret bounds relies on a discretization of the space.
Morever, once $\hat{m}$ is chosen, the acquisition function can be optimized with any search method, including gradient descent. 
\paragraph{High dimensions.}
\label{ssec:dim}
Bayesian Optimization methods generally suffer in high dimensions. Common assumptions are a low-dimensional or simpler underlying structure of~$f$  \cite{djolonga2013high,wang2013,kandasamy2015high}. Our approach can be combined with those methods too, to be extended to higher dimensions.
\paragraph{Relation to entropy search. }
EST is closely related to
entropy search (ES) methods~\citep{hennig2012,hernandez2014predictive}, but also differs in a few important aspects. Like EST, ES methods approximate the probability of a point $\vx$ being the maximizer $\argmax_{\vx'\in\mathfrak X } f(\vx')$, and then choose where to evaluate next by optimizing an acquisition function related to this probability. %
However, instead of choosing the input that is most likely to be the $\argmax$, ES chooses where to evaluate next by optimizing the expected change in the entropy of $\Pr[M_\vx]$. %
One reason is that ES does not aim to minimize cumulative regret like many other bandit methods (including EST). The cumulative regret penalizes all queried points, and a method that minimizes the cumulative regret needs to query enough supposedly good points. ES methods, in contrast, purely focus on exploration, since their objective is to gather as much information as possible to  estimate a final value in the very end. %
Since the focus of this work lies on cumulative regret, detailed empirical comparisons between EST and ES may be found in the supplement.

\section{Conclusion}
In this paper, we studied a new Bayesian optimization strategy derived from the viewpoint of the estimating the $\argmax$ of an unknown function. We showed that this strategy corresponds to adaptively setting the trade-off parameters $\lambda$ and $\theta$ in GP-UCB and GP-PI, and established bounds on the regret. Our experiments demonstrate that this strategy is not only easy to use, but robustly performs well by measure of different types of regret, on a variety of real-world tasks from robotics and computer vision.

\paragraph{Acknowledgements.}
We thank Leslie Kaelbling and Tom\'as Lozano-P\'erez for discussions, and Antonio Torralba for support with computational resources.
We gratefully acknowledge support from NSF CAREER award 1553284, NSF grants 1420927 and 1523767, from ONR grant N00014-14-1-0486, and from ARO grant W911NF1410433.  Any opinions, findings, and conclusions or recommendations expressed in this material are those of the authors and do not necessarily reflect the views of our sponsors.

{\small
\bibliographystyle{abbrvnat}
\bibliography{refs}

\begin{thebibliography}{37}
\providecommand{\natexlab}[1]{#1}
\providecommand{\url}[1]{\texttt{#1}}
\expandafter\ifx\csname urlstyle\endcsname\relax
  \providecommand{\doi}[1]{doi: #1}\else
  \providecommand{\doi}{doi: \begingroup \urlstyle{rm}\Url}\fi

\bibitem[Auer(2002)]{auer2002b}
P.~Auer.
\newblock Using confidence bounds for exploitation-exploration tradeoffs.
\newblock \emph{Journal of Machine Learning Research}, 3:\penalty0 397--422,
  2002.

\bibitem[Auer et~al.(2002)Auer, Cesa-Bianchi, and Fischer]{auer2002finite}
P.~Auer, N.~Cesa-Bianchi, and P.~Fischer.
\newblock Finite-time analysis of the multiarmed bandit problem.
\newblock \emph{Machine learning}, 47\penalty0 (2-3):\penalty0 235--256, 2002.

\bibitem[Brochu et~al.(2009)Brochu, Cora, and De~Freitas]{brochu2009}
E.~Brochu, V.~M. Cora, and N.~De~Freitas.
\newblock A tutorial on {B}ayesian optimization of expensive cost functions,
  with application to active user modeling and hierarchical reinforcement
  learning.
\newblock Technical Report TR-2009-023, University of British Columbia, 2009.

\bibitem[Bubeck et~al.(2009)Bubeck, Munos, and Stoltz]{bubeck2009pure}
S.~Bubeck, R.~Munos, and G.~Stoltz.
\newblock Pure exploration in multi-armed bandits problems.
\newblock In \emph{Algorithmic Learning Theory}, pages 23--37. Springer, 2009.

\bibitem[Bull(2011)]{bull2011}
A.~D. Bull.
\newblock Convergence rates of efficient global optimization algorithms.
\newblock \emph{Journal of Machine Learning Research}, 12:\penalty0 2879--2904,
  2011.

\bibitem[Calandra et~al.(2014)Calandra, Seyfarth, Peters, and
  Deisenroth]{calandra2014experimental}
R.~Calandra, A.~Seyfarth, J.~Peters, and M.~P. Deisenroth.
\newblock An experimental comparison of {B}ayesian optimization for bipedal
  locomotion.
\newblock In \emph{International Conference on Robotics and Automation (ICRA)},
  2014.

\bibitem[Djolonga et~al.(2013)Djolonga, Krause, and Cevher]{djolonga2013high}
J.~Djolonga, A.~Krause, and V.~Cevher.
\newblock High-dimensional {G}aussian process bandits.
\newblock In \emph{Advances in Neural Information Processing Systems (NIPS)},
  2013.

\bibitem[Fei-Fei et~al.(2007)Fei-Fei, Fergus, and Perona]{fei2007learning}
L.~Fei-Fei, R.~Fergus, and P.~Perona.
\newblock Learning generative visual models from few training examples: An
  incremental {B}ayesian approach tested on 101 object categories.
\newblock \emph{Computer Vision and Image Understanding}, 2007.

\bibitem[Gill et~al.(2002)Gill, Murray, and Saunders]{gill2002snopt}
P.~E. Gill, W.~Murray, and M.~A. Saunders.
\newblock {SNOPT}: An {SQP} algorithm for large-scale constrained optimization.
\newblock \emph{SIAM Journal on optimization}, 12\penalty0 (4):\penalty0
  979--1006, 2002.

\bibitem[Griffin et~al.(2007)Griffin, Holub, and Perona]{griffin2007caltech}
G.~Griffin, A.~Holub, and P.~Perona.
\newblock Caltech-256 object category dataset.
\newblock Technical Report 7694, California Institute of Technology, 2007.

\bibitem[Gr{\"u}new{\"a}lder et~al.(2010)Gr{\"u}new{\"a}lder, Audibert, Opper,
  and Shawe-Taylor]{grunewalder2010regret}
S.~Gr{\"u}new{\"a}lder, J.-Y. Audibert, M.~Opper, and J.~Shawe-Taylor.
\newblock Regret bounds for {G}aussian process bandit problems.
\newblock In \emph{International Conference on Artificial Intelligence and
  Statistics (AISTATS)}, 2010.

\bibitem[Hennig and Schuler(2012)]{hennig2012}
P.~Hennig and C.~J. Schuler.
\newblock Entropy search for information-efficient global optimization.
\newblock \emph{Journal of Machine Learning Research}, 13:\penalty0 1809--1837,
  2012.

\bibitem[Hern{\'a}ndez-Lobato et~al.(2014)Hern{\'a}ndez-Lobato, Hoffman, and
  Ghahramani]{hernandez2014predictive}
J.~M. Hern{\'a}ndez-Lobato, M.~W. Hoffman, and Z.~Ghahramani.
\newblock Predictive entropy search for efficient global optimization of
  black-box functions.
\newblock In \emph{Advances in Neural Information Processing Systems (NIPS)},
  2014.

\bibitem[Jia(2013)]{Jia13caffe}
Y.~Jia.
\newblock {Caffe}: An open source convolutional architecture for fast feature
  embedding.
\newblock \url{http://caffe.berkeleyvision.org/}, 2013.

\bibitem[Kandasamy et~al.(2015)Kandasamy, Schneider, and
  Poczos]{kandasamy2015high}
K.~Kandasamy, J.~Schneider, and B.~Poczos.
\newblock High dimensional {B}ayesian optimisation and bandits via additive
  models.
\newblock In \emph{International Conference on Machine Learning (ICML)}, 2015.

\bibitem[Krause and Ong(2011)]{krause2011contextual}
A.~Krause and C.~S. Ong.
\newblock Contextual {G}aussian process bandit optimization.
\newblock In \emph{Advances in Neural Information Processing Systems (NIPS)},
  2011.

\bibitem[Kushner(1964)]{kushner1964}
H.~J. Kushner.
\newblock A new method of locating the maximum point of an arbitrary multipeak
  curve in the presence of noise.
\newblock \emph{Journal of Fluids Engineering}, 86\penalty0 (1):\penalty0
  97--106, 1964.

\bibitem[Li and Fei-Fei(2007)]{li2007and}
L.-J. Li and L.~Fei-Fei.
\newblock What, where and who? classifying events by scene and object
  recognition.
\newblock In \emph{International Conference on Computer Vision (ICCV)}, 2007.

\bibitem[Lizotte et~al.(2007)Lizotte, Wang, Bowling, and
  Schuurmans]{lizotte2007}
D.~J. Lizotte, T.~Wang, M.~H. Bowling, and D.~Schuurmans.
\newblock Automatic gait optimization with {G}aussian process regression.
\newblock In \emph{International Conference on Artificial Intelligence
  (IJCAI)}, 2007.

\bibitem[Lizotte et~al.(2012)Lizotte, Greiner, and Schuurmans]{lizotte2012}
D.~J. Lizotte, R.~Greiner, and D.~Schuurmans.
\newblock An experimental methodology for response surface optimization
  methods.
\newblock \emph{Journal of Global Optimization}, 53\penalty0 (4):\penalty0
  699--736, 2012.

\bibitem[Massart(2007)]{massart2007concentration}
P.~Massart.
\newblock \emph{Concentration Inequalities and Model Selection}, volume~6.
\newblock Springer, 2007.

\bibitem[Mo{\u{c}}kus(1974)]{mockus1974}
J.~Mo{\u{c}}kus.
\newblock On {B}ayesian methods for seeking the extremum.
\newblock In \emph{Optimization Techniques IFIP Technical Conference}, 1974.

\bibitem[Quattoni and Torralba(2009)]{quattoni2009recognizing}
A.~Quattoni and A.~Torralba.
\newblock Recognizing indoor scenes.
\newblock In \emph{IEEE Conference on Computer Vision and Pattern Recognition
  (CVPR)}, 2009.

\bibitem[Rasmussen and Williams(2006)]{rasmussen2006gaussian}
C.~E. Rasmussen and C.~K. Williams.
\newblock Gaussian processes for machine learning.
\newblock \emph{The MIT Press}, 2006.

\bibitem[Razavian et~al.(2014)Razavian, Azizpour, Sullivan, and
  Carlsson]{razavian2014cnn}
A.~S. Razavian, H.~Azizpour, J.~Sullivan, and S.~Carlsson.
\newblock {CNN} features off-the-shelf: an astounding baseline for recognition.
\newblock In \emph{IEEE Conference on Computer Vision and Pattern Recognition
  (CVPR)}, 2014.

\bibitem[Ross(2003)]{ross2003useful}
A.~M. Ross.
\newblock Useful bounds on the expected maximum of correlated normal variables.
\newblock Technical report, Technical Report 03W-004, ISE Dept., Lehigh Univ.,
  Aug, 2003.

\bibitem[Schulman et~al.(2013)Schulman, Ho, Lee, Awwal, Bradlow, and
  Abbeel]{schulman2013finding}
J.~Schulman, J.~Ho, A.~Lee, I.~Awwal, H.~Bradlow, and P.~Abbeel.
\newblock Finding locally optimal, collision-free trajectories with sequential
  convex optimization.
\newblock In \emph{Robotics: Science and Systems Conference (RSS)}, volume~9,
  pages 1--10, 2013.

\bibitem[Slepian(1962)]{slepian1962one}
D.~Slepian.
\newblock The one-sided barrier problem for {G}aussian noise.
\newblock \emph{Bell System Technical Journal}, 41\penalty0 (2):\penalty0
  463--501, 1962.

\bibitem[Snoek et~al.(2012)Snoek, Larochelle, and Adams]{snoek2012practical}
J.~Snoek, H.~Larochelle, and R.~P. Adams.
\newblock Practical {B}ayesian optimization of machine learning algorithms.
\newblock In \emph{Advances in Neural Information Processing Systems (NIPS)},
  2012.

\bibitem[Srinivas et~al.(2010)Srinivas, Krause, Kakade, and
  Seeger]{srinivas2009gaussian}
N.~Srinivas, A.~Krause, S.~M. Kakade, and M.~Seeger.
\newblock Gaussian process optimization in the bandit setting: No regret and
  experimental design.
\newblock In \emph{International Conference on Machine Learning (ICML)}, 2010.

\bibitem[Tedrake()]{tedrake14}
R.~Tedrake.
\newblock \emph{Underactuated Robotics: Algorithms for Walking, Running,
  Swimming, Flying, and Manipulation (Course Notes for MIT 6.832)}.
\newblock Downloaded in Fall, 2014 from
  \url{http://people.csail.mit.edu/russt/underactuated/}.

\bibitem[Tedrake(2014)]{drake}
R.~Tedrake.
\newblock Drake: A planning, control, and analysis toolbox for nonlinear
  dynamical systems.
\newblock \url{http://drake.mit.edu}, 2014.

\bibitem[Thornton et~al.(2013)Thornton, Hutter, Hoos, and
  Leyton-Brown]{thornton13}
C.~Thornton, F.~Hutter, H.~H. Hoos, and K.~Leyton-Brown.
\newblock Auto-{WEKA}: combined selection and hyperparameter optimization of
  classification algorithms.
\newblock In \emph{ACM SIGKDD Conference on Knowledge Discovery and Data Mining
  (KDD)}, 2013.

\bibitem[Wang et~al.(2013)Wang, Zoghi, Hutter, Matheson, and
  De~Freitas]{wang2013}
Z.~Wang, M.~Zoghi, F.~Hutter, D.~Matheson, and N.~De~Freitas.
\newblock Bayesian optimization in high dimensions via random embeddings.
\newblock In \emph{International Conference on Artificial Intelligence
  (IJCAI)}, 2013.

\bibitem[Xiao et~al.(2010)Xiao, Hays, Ehinger, Oliva, and
  Torralba]{xiao2010sun}
J.~Xiao, J.~Hays, K.~A. Ehinger, A.~Oliva, and A.~Torralba.
\newblock Sun database: large-scale scene recognition from abbey to zoo.
\newblock In \emph{IEEE Conference on Computer Vision and Pattern Recognition
  (CVPR)}, 2010.

\bibitem[Yao et~al.(2011)Yao, Jiang, Khosla, Lin, Guibas, and
  Fei-Fei]{yao2011human}
B.~Yao, X.~Jiang, A.~Khosla, A.~L. Lin, L.~Guibas, and L.~Fei-Fei.
\newblock Human action recognition by learning bases of action attributes and
  parts.
\newblock In \emph{International Conference on Computer Vision (ICCV)}, 2011.

\bibitem[Zhou et~al.(2014)Zhou, Lapedriza, Xiao, Torralba, and
  Oliva]{zhou2014learning}
B.~Zhou, A.~Lapedriza, J.~Xiao, A.~Torralba, and A.~Oliva.
\newblock Learning deep features for scene recognition using places database.
\newblock In \emph{Advances in Neural Information Processing Systems (NIPS)},
  2014.

\end{thebibliography}
}
\vspace{1.5em}
\newpage
\appendix
\twocolumn[

\center{
\Large\textbf{Supplement}
}
\vspace{1.5em}
]

In this supplement, we provide proofs for all theorems and lemmas in the main paper, more exhaustive experimental results and details on the experiments.

\section{Proofs}

\subsection{Proofs from Section 2}
\setcounter{section}{2}
\setcounter{thm}{0}
\ucblemma*
\begin{proof}
We omit the subscripts $t$ for simplicity. Let $\va$ be the point selected by GP-UCB, and $\vb$ selected by EST. Without loss of generality, we assume $\va$ and $\vb$ are unique. 
With $\lambda = \min_{\vx\in\mathfrak X} \frac{\hat{m}-\mu(\vx)}{\sigma(\vx)}$, GP-UCB chooses to evaluate 
$$\va =\argmax_{\vx\in\mathfrak X}\mu(\vx) + \lambda\sigma(\vx)= \argmin_{\vx\in\mathfrak X} \frac{\hat{m}-\mu(\vx)}{\sigma(\vx)}.$$
This is because $$\hat m = \max_{\vx\in\mathfrak X} \mu(\vx) + \lambda\sigma(\vx)= \mu(\va) + \lambda\sigma(\va).$$
By definition of $\vb$, for all $\vx\in\mathfrak{X}$,  we have
\begin{align}
\frac{\Pr[M_{\vb} | \hat m,\mathfrak D]} {\Pr[M_\vx| \hat m,\mathfrak D]} &\approx%
 \frac{Q(\frac{\hat{m}-\mu(\vb)}{\sigma(\vb)})\prod_{\vx'\neq \vb}\Phi(\frac{\hat{m}-\mu(\vx')}{\sigma(\vx')})}{Q(\frac{\hat{m}-\mu(\vx)}{\sigma(\vx)})\prod_{\vx'\neq \vx}\Phi(\frac{\hat{m}-\mu(\vx')}{\sigma(\vx')})} \nonumber\\
 &=
  \frac{Q(\frac{\hat{m}-\mu(\vb)}{\sigma(\vb)}) \Phi(\frac{\hat{m}-\mu(\vx)}{\sigma(\vx)})}{Q(\frac{\hat{m}-\mu(\vx)}{\sigma(\vx)})\Phi(\frac{\hat{m}-\mu(\vb)}{\sigma(\vb)})} \nonumber\\
  &\geq 1 .\nonumber
\end{align}
The inequality holds if and only if $\frac{\hat{m}-\mu(\vb)}{\sigma(\vb)}\leq \frac{\hat{m}-\mu(\vx)}{\sigma(\vx)}$ for all $\vx\in\mathfrak X$, including $\va$, and hence
\begin{align}
\frac{\hat{m}-\mu(\vb)}{\sigma(\vb)}\leq \frac{\hat{m}-\mu(\va)}{\sigma(\va)} = \lambda = \min_{\vx\in\mathfrak X} \frac{\hat{m}-\mu (\vx)}{\sigma (\vx)} ,\nonumber %
\end{align}
which, with uniqueness, implies that $\va=\vb$ and GP-UCB and EST select the same point.

For the other direction, we denote the candidate selected by GP-UCB by $$\va = \argmax_{\vx\in\mathfrak X}{\mu(\vx) + \lambda\sigma(\vx)}.$$
 The variant of EST with $\hat m = \max_{\vx\in\mathfrak X}{\mu (\vx) + \lambda\sigma(\vx)}$ selects $$\vb= \argmax_{\vx\in\mathfrak X}\Pr[M_{\vx} | \hat m,\mathfrak D].$$
We know that for all $\vx\in\mathfrak X$, we have $\frac{\hat m-\mu(\vb)}{\sigma(\vb)}\leq \frac{\hat m-\mu(\vx)}{\sigma(\vx)}$ and hence $\hat m\leq \mu(\vb) + \frac{\hat m-\mu(\vx)}{\sigma(\vx)} \sigma(\vb)$.
Since $\hat m =\mu(\va) + \lambda\sigma(\va) $, letting $\vx = \va$ implies that
\begin{align}
\hat m = \max_{\vx\in\mathfrak X}{\mu (\vx) + \lambda\sigma(\vx)} \leq \mu(\vb) + \lambda \sigma(\vb) .\nonumber%
\end{align}
Hence, by uniqueness it must be that $\va=\vb$ and GP-UCB and EST select the same candidate.
\end{proof}

\hide{
\begin{proof}
We omit the subscripts $t$ for simplicity. With $\beta^{\frac12} = \min_{\vx\in\mathfrak X} \frac{m-\mu(\vx)}{\sigma(\vx)}$, UCB plays $\va = \argmin_{\vx\in\mathfrak X} \frac{m-\mu(\vx)}{\sigma(\vx)}$. Let $\vb$ be the candidate selected by EST. Without loss of generality, we assume $\va$ and $\vb$ are unique (otherwise use a small perturbation). By definition of $\vb$, for all $\vx\in\mathfrak{X}$,  we have
\begin{align}
1 &\;\leq\; \frac{\Pr[M_{\vb} | m,\mathfrak D]} {\Pr[M_\vx| m,\mathfrak D]} \\
&=
 \frac{Q(\frac{m-\mu(\vb)}{\sigma(\vb)})\prod_{\vx'\neq \vb}\Phi(\frac{m-\mu(\vx')}{\sigma(\vx')})}{Q(\frac{m-\mu(\vx)}{\sigma(\vx)})\prod_{\vx'\neq \vx}\Phi(\frac{m-\mu(\vx')}{\sigma(\vx')})}\\
 &=
  \frac{Q(\frac{m-\mu(\vb)}{\sigma(\vb)}) \Phi(\frac{m-\mu(\vx)}{\sigma(\vx)})}{Q(\frac{m-\mu(\vx)}{\sigma(\vx)})\Phi(\frac{m-\mu(\vb)}{\sigma(\vb)})}. \label{prmneq}
\end{align}
The inequality holds if and only if it holds that $\frac{m-\mu(\vb)}{\sigma(\vb)}\leq \frac{m-\mu(\vx)}{\sigma(\vx)}$ for all $\vx\in\mathfrak X$, including $\va$, and hence
\begin{align}
\frac{m-\mu(\vb)}{\sigma(\vb)}\leq \frac{m-\mu(\va)}{\sigma(\va)} = \beta^{\frac12} = \min_{\vx\in\mathfrak X} \frac{m-\mu (\vx)}{\sigma (\vx)} \label{thm2contra}
\end{align}
which, with uniqueness, implies that $\va=\vb$ and UCB and EST select the same candidate.

For the other direction, denote the candidate selected by UCB by $\va = \argmax_{\vx\in\mathfrak X}{\mu(\vx) + \beta^{\frac12}\sigma(\vx)}$. The variant of EST with $m = \max_{\vx\in\mathfrak X}{\mu (\vx) + \beta^{\frac12}\sigma(\vx)}$ selects $\vb= \argmax_{\vx\in\mathfrak X}\Pr[M_{\vx} | m,\mathfrak D]$.
We know that for all $\vx\in\mathfrak X$, we have $\frac{m-\mu(\vb)}{\sigma(\vb)}\leq \frac{m-\mu(\vx)}{\sigma(\vx)}$ and hence $m\leq \mu(\vb) + \frac{m-\mu(\vx)}{\sigma(\vx)} \sigma(\vb)$.
Since $m =\mu(\va) + \beta^{\frac12}\sigma(\va) $, letting $\vx = \va$ implies that
\begin{align}
m = \max_{\vx\in\mathfrak X}{\mu (\vx) + \beta^{\frac12}\sigma(\vx)} \leq \mu(\vb) + \beta^{\frac12} \sigma(\vb) \label{thm1contra}.
\end{align}
Hence, by uniqueness it must be that $\va=\vb$ and UCB and EST select the same candidate.
\end{proof}}
\setcounter{section}{1}
\subsection{Proofs from Section 3}
\setcounter{section}{3}
\setcounter{thm}{1}

\pbound*
\begin{proof}
Let $z_t=\frac{\mu_{t-1}(x_t)-f(x_t)}{\sigma_{t-1}(x_t)}\sim\mathcal N(0,1) $. It holds that %
\begin{align*}
\Pr[z_t>\zeta_t ] &= \int_{\zeta_t }^{+\infty} \frac{1}{\sqrt{2\pi}}e^{-z^2/2}\dif z \\
&=\int_{\zeta_t }^{+\infty} \frac{1}{\sqrt{2\pi}}e^{-(z-\zeta_t )^2/2-\zeta_t^2/2-z\zeta_t }\dif z\\
&\leq e^{-\zeta_t^2/2}\int_{\zeta_t }^{+\infty} \frac{1}{\sqrt{2\pi}}e^{-(z-\zeta_t )^2/2}\dif z\\
&=\frac12 e^{-\zeta_t^2/2}.
\end{align*}
A union bound extends this bound to all rounds:
\begin{align*}
\Pr[z_t>\zeta_t  \text{ for some } t\in [1,T]] \leq \sum_{t=1}^T\frac{1}{2}e^{-\zeta_t^2/2}.
\end{align*}
With $\zeta_t=(2\log(\frac{\pi_t}{2\delta}))^\frac12$ and $\sum_{t=1}^T \pi_t^{-1} = 1$, this implies that with probability at least $1-\delta$, it holds that $\mu_{t-1}(\vx_{t}) -f( \vx_{t})  \leq \zeta_t\sigma_{t-1}(\vx_{t})$ for all $t \in [1,T]$. One may set $\pi_t = \tfrac16 \pi^2 t^2$, or $\pi_t = T$, in which case $\zeta_t = \zeta = (2\log(\frac{T}{2\delta}))^\frac12$.
\end{proof}

\regretlem*
\begin{proof}
  At time step $t\geq 1$, we have
\begin{align}
\rt_t & = \max_{\vx\in\mathfrak X}f(\vx)-  f(\vx_{t})\nonumber \\
&\leq \hat m_t - f(\vx_{t}) \nonumber\\
&\leq \hat m_t - \mu_{t-1}(\vx_t) + \zeta_t \sigma_{t-1}(\vx_{t}) \nonumber\\
&= (\bt_t +  \zeta_t  ) \sigma_{t-1}(\vx_{t}). \nonumber
\end{align}
\end{proof}

\hide{
\begin{lem}
\label{expdlem}
For EST, the expected regret at time step $t$ is $\mathbb{E}[\rt_t] = \bt_t^{\frac12} \sigma_{t-1}(\vx_{t})$. The expected simple regret is upper bounded by $\mathbb{E}[r_t] = \mathbb{E}[\min_{1\leq t \leq T} \rt_t] \leq \bt_T^{\frac12} \sigma_{T-1}(\vx_{T})$.
\end{lem}

\begin{proof}
At time step $t\geq 1$, we have
\begin{align*}
\Ex[\rt_t] &= \Ex[Y- f(\vx_t) | \mathfrak D] = m - \mu_{t-1}(\vx_{t}) = \bt_t^{\frac12} \sigma_{t-1}(\vx_{t}).
\end{align*}
From this, the expected simple regret can be bounded via Jensen's inequality:
\begin{align*}
\Ex[\min_{1 \leq t \leq T}{\rt_t}] \leq \min_t\Ex[\rt_t] \leq \Ex[\rt_T]  = \bt_T^{\frac12} \sigma_{T-1}(\vx_{T})
\end{align*}
\end{proof}}

\setcounter{section}{1}
\section{Experiments}
\subsection{Approximate $m$}
\begin{figure}
        \centering
        \includegraphics[width=0.45\textwidth]{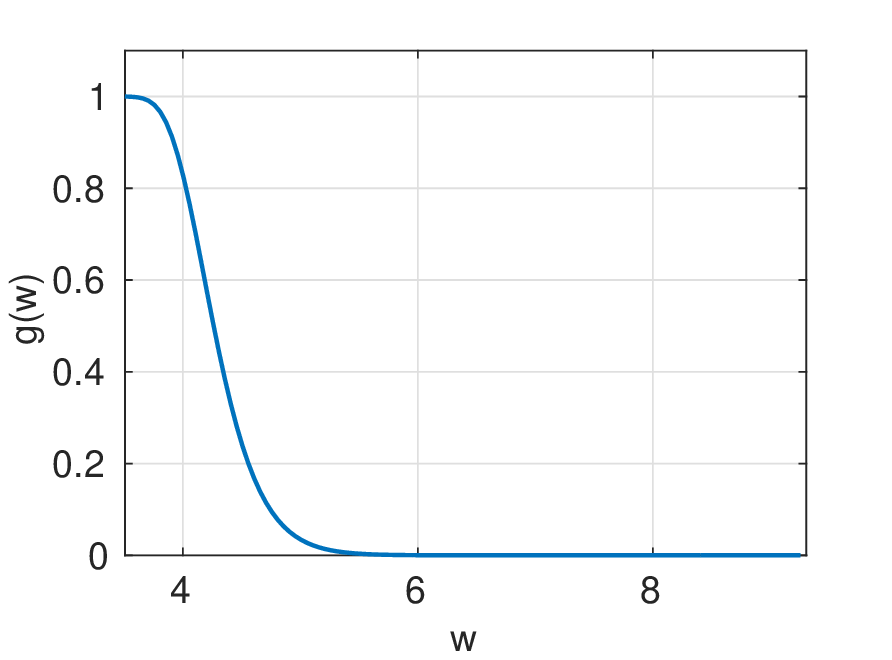}
        \hspace{20pt}
        \includegraphics[width=0.45\textwidth]{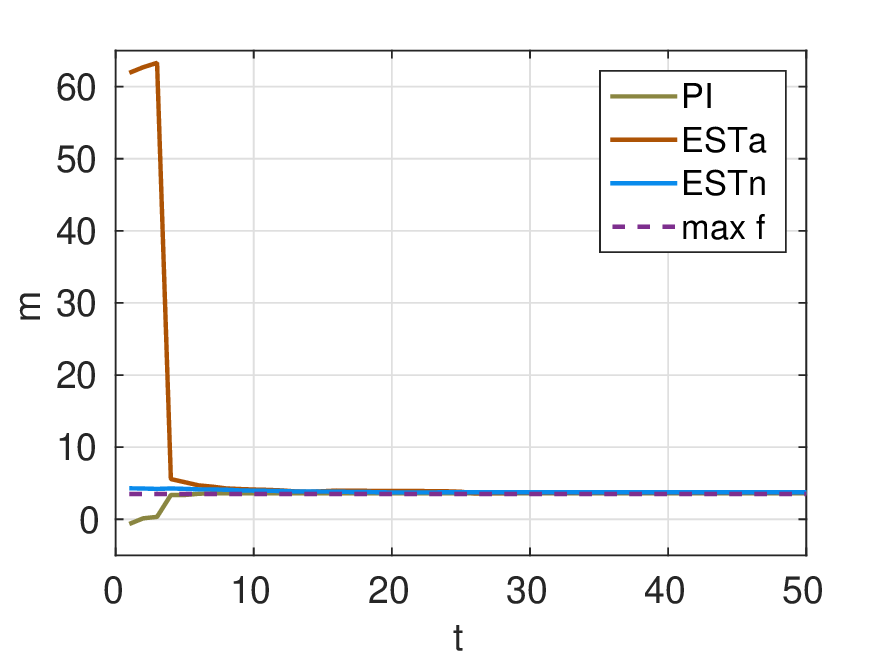}
        \caption{Top: $g(w)$, $w \in [m_0,+\infty)$; Bottom: estimation of $m$.}
        \label{fig:estm}
\end{figure}

In the paper, we estimate $m$ to be 

\begin{align*}
\hat{m} = m_0+\int_{m_0}^\infty 1- \prod_{\vx\in\mathfrak W} \Phi\Big(\frac{w-\mu(\vx)}{\sigma(\vx)}\Big) \dif w %
\end{align*}
which involves an integration from the current maximum $m_0$ of the observed data to positive infinity. In fact the factor inside the integration quickly approaches zero in practice. We plot $g(w) = 1- \prod_{\vx\in\mathfrak W} \Phi\Big(\frac{w-\mu(\vx)}{\sigma(\vx)}\Big)$ in Figure~\ref{fig:estm}, which looks like half of a Gaussian distribution. So instead of numerical integration (which can be done efficiently), heuristically we can sample two values of $g(w)$ to fit $\hat g(w) = a e^{-(w-m_0)^2/2b^2}$ and do the integration $\int_{m_0}^\infty \hat g(w) \dif w = \sqrt{2\pi}ab$ analytically to be more efficient. This method is what we called ESTa in the paper, while the numerical integration is called ESTn. 

We notice that our estimation $\hat m$ can serve as a tight upper bound on the real value of the max of the function in practice. %
One example of is shown in Figure~\ref{fig:estm} with a 1-D GP function. This example shows how \PI, \ESTa\ and \ESTn\ estimate $m$. Both \ESTa\ and \ESTn\ are upper bounds of the true maximum of the function, and \ESTn\ is actually very tight. For \PI, $\theta = \argmax_{1\leq \tau<t} y_\tau + \epsilon$ is always a lower bound of an $\epsilon$ shift over the true maximum of the function.
\begin{figure}
        \centering
        \includegraphics[width=0.45\textwidth]{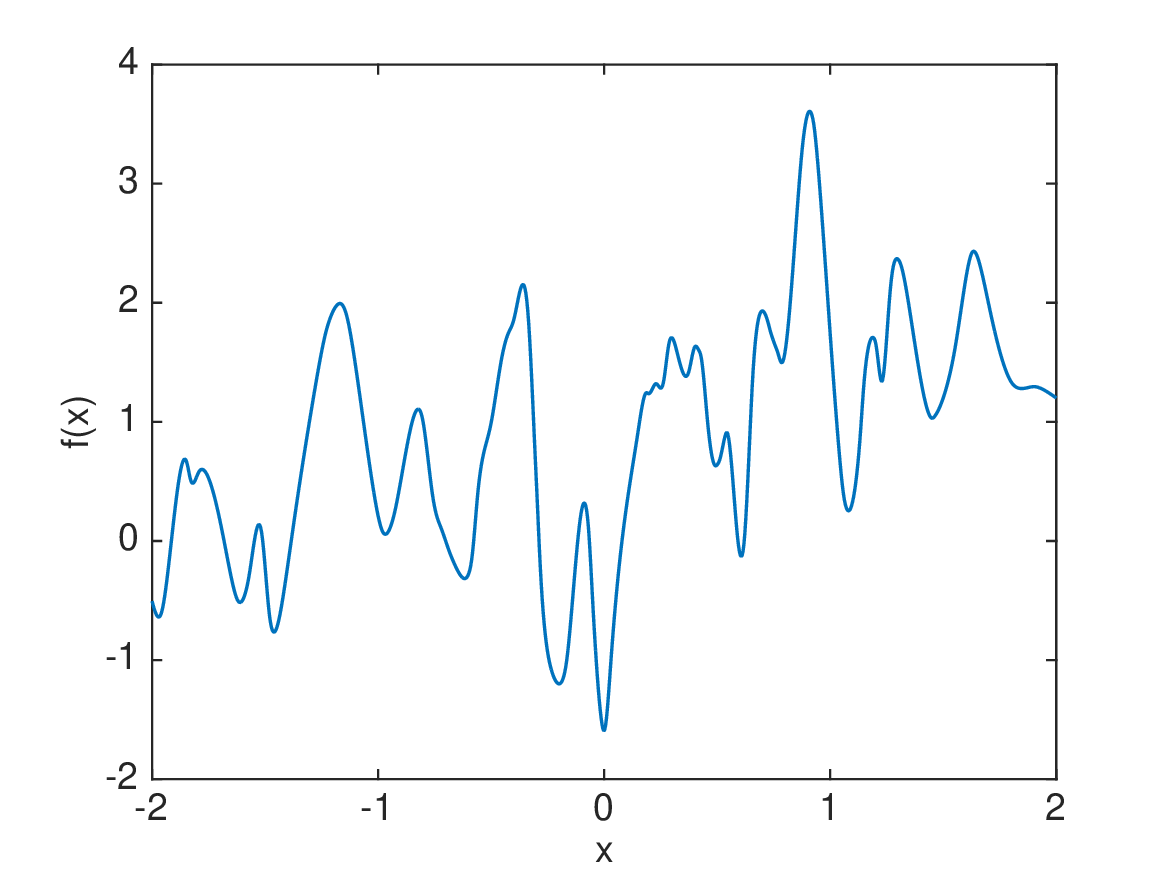}
        \hspace{20pt}
        \includegraphics[width=0.45\textwidth]{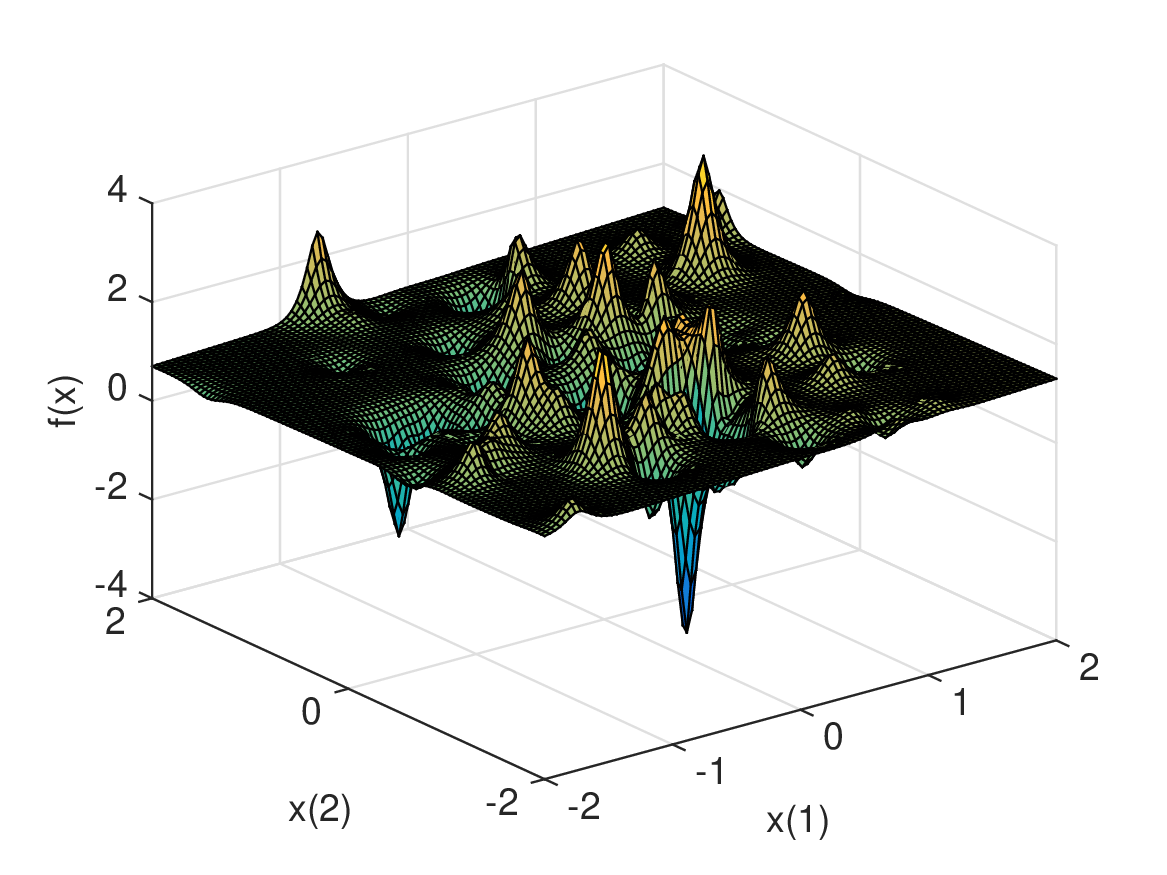}
        \caption{Examples of a function sampled from 1-D GP (top), and a function sampled from 2-D GP (bottom) with isotropic Mat\'ern kernel and linear mean function. We deliberately create many local optimums to make the problem hard.}
        \label{fig:gpexample}
\end{figure}

\subsection{Synthetic data}
\label{sec:supsynth}
We show the examples of the functions we sampled from GP in Figure~\ref{fig:gpexample}. The covariance function of GP is an isotropic Mat\'ern kernel with parameters $\ell = 0.1, \sigma_f = 1$.
The mean function is a linear function with a fixed random slope for different dimensions, and the constant is $1$.

\subsection{Initialization tuning for trajectory optimization}
The 8 configurations of start state are $[7\; 1\; 0\; 0]$, $[7\; 0\; 0\; 0]$, $[1\; 0\; 0\; 0]$, $[1\; 1\; 0\; 0]$, $[2\; 0\; 0\; 0]$, $[2\; 1\; 0\; 0]$, $[3\; 0\; 0\; 0]$, $[3\; 1\; 0\; 0]$, where the first two dimensions denote the position and the last two dimension denote the speed. We only tune the first two dimension and keep the speed to be 0 for both directions. The target state is fixed to be $[5\;9\;0\;0]$.

We can initialize the trajectory by setting the mid point of trajectory to be any point falling on the grid of the space (both x axis and y axis have range $[-2,12] $). Then use SNOPT to solve the trajectory optimization problem, which involves an objective cost function (we take the negative cost to be a reward function to maximize), dynamics constraints, and obstacle constraints etc. Details of trajectory optimization are available in \cite{drake,tedrake14}.

We used the same settings of parameters for GP as in Section~\ref{sec:supsynth} for all the methods we tested and did kernel parameter fitting every 5 rounds. The same strategy was used for the image classification experiments in the next section.
\subsection{Parameter tuning for image classification}

We use the linear SVM in the liblinear package for all the image classification experiments.  We extract the FC7 activation from the imagenet reference network in the Caffe deep learning package~\cite{Jia13caffe} as the visual feature. The reported classification accuracy is the accuracy averaged over all the categories. `-c' cost is the model parameter we tune for the linear SVM.

In Caltech101 and Caltech256 experiment~\cite{fei2007learning,griffin2007caltech}, there are 8,677 images from 101 object categories in the Caltech101 and 29,780 images from 256 object categories. The training size is 30 images per category, and the rest are test images. 

In SUN397 experiment~\cite{xiao2010sun}, there are 108,754 images from 397 scene categories. Images are randomly split into training and test set. The training size is 50 images per category, and the rest are test images. 

In MIT Indoor67 experiment~\cite{quattoni2009recognizing}, there are 15,620 images from 67 indoor scene categories. Images are randomly split into training set and test set. The training size is 100 images per category, and the rest are test images. 

In Stanford Action40 experiment~\cite{yao2011human}, there are 9,532 images from 40 action categories. Images are randomly split into training set and test set. The training size is 100 images per category, and the rest are test images. 

In UIUC Event8 experiment~\cite{li2007and}, there are 1,579 images from 8 event categories. Images are randomly split into training set and test set.  Training size is 70 images per category,  and the rest are test images. 

\begin{sidewaysfigure*}
\begin{center}
\includegraphics[width=1\linewidth]{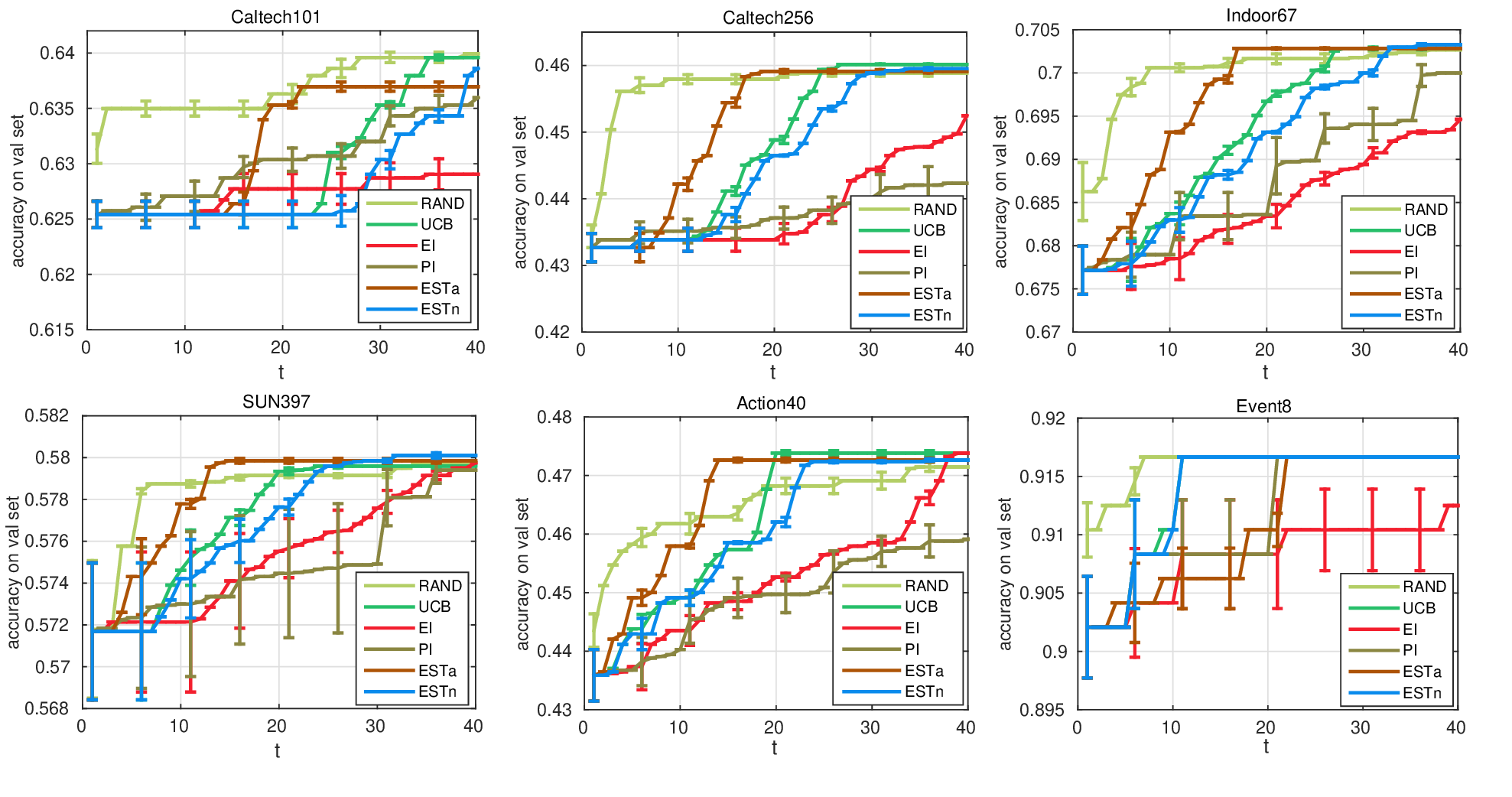}
\end{center}
   \caption{Maximum accuracy on the validation set over iteration of the optimization. %
   Experiments are repeated 5 times. The visual features used here are Deep CNN features pre-trained on the Places database.}\label{ponvalidationset}
\end{sidewaysfigure*}

We used features extracted from a convolutional neural network (CNN) that was trained on images from ImageNet.  It has been found~\cite{zhou2014learning} that features from a CNN trained on a set of images focused more on places than on objects, the Places database, work better in some domains.  So, we repeated our experiments using the Places-CNN features, and the results are shown in Figure~\ref{ponvalidationset}  and Table~\ref{pontestset}.  All the methods help to improve the classification accuracy on the validation set. EST methods achieve good accuracy on each validation set on par with the best competitors for most of the datasets. And we also observe that for Caltech101 and Event8, \RND\ and \UCB\ converge faster and achieve better accuracy than other methods. As we have shown in Section~\ref{sec:synthexp}, \UCB\ and \RND\ perform worse than other methods in terms of cumulative regret, because they tend to explore too much. However, more exploration can be helpful for some black-box functions that do not satisfy our assumption that they are samples from GP. For example, for discontinuous step functions, pure exploration can be beneficial for simple regret. One possible explanation for the better results of \RND\ and \UCB\ is that the black-box functions we optimize here are possibly functions not satisfying our assumption. The strong assumption on the black-box function is also a major drawback of Bayesian optimization,  %

\begin{sidewaystable*}\caption{Classification accuracy on the test set of the datasets after the model parameter is tuned by ESTa and ESTn. Tuning achieves good improvement over the results in~\cite{zhou2014learning}.}\label{pontestset}
\centering
\begin{tabular}{|c|c|c|c|c|c|c|}
\hline
& Caltech101 & Caltech256 & Indoor67 & SUN397 & Action40 & Event8 \\
\hline
\hline
Imagenet-CNN feature & 87.22 & 67.23 & 56.79 & 42.61 & 54.92 & 94.42 \\
\hline
ESTa & 88.23 & 69.39 & 60.02 & 47.13 & 57.60 & 94.91\\
ESTn & 88.25 & 69.39 & 60.08 & 47.21 & 57.58 & 94.86\\
\hline
\hline
Places-CNN feature& 65.18 & 45.59 & 68.24 & 54.32 & 42.86 & 94.12 \\
\hline
ESTa & 66.94 & 47.51 & 70.27 & 58.57 & 46.24 & 93.79\\
ESTn & 66.95 & 47.43 & 70.27 & 58.65 & 46.17 & 93.56\\

\hline
\end{tabular}

\end{sidewaystable*}

\subsection{Comparison to entropy search methods}

Entropy search methods~\cite{hennig2012,hernandez2014predictive} aim to minimize the entropy of the probability for the event $M_{\vx}$ ($\vx = \argmax_{\vx'\in\mathfrak X} f(\vx')$). Although not suitable for minimizing cumulative regret, ES methods are intuitively ideal for minimizing simple regret. We hence in this section compare the empirical performance of entropy search (ES)~\cite{hennig2012} and predictive entropy search (PES)~\cite{hernandez2014predictive} to that of the EST methods (EST/GP-UCB/PI) and EI.

\begin{figure}
        \centering
        \includegraphics[width=0.45\textwidth]{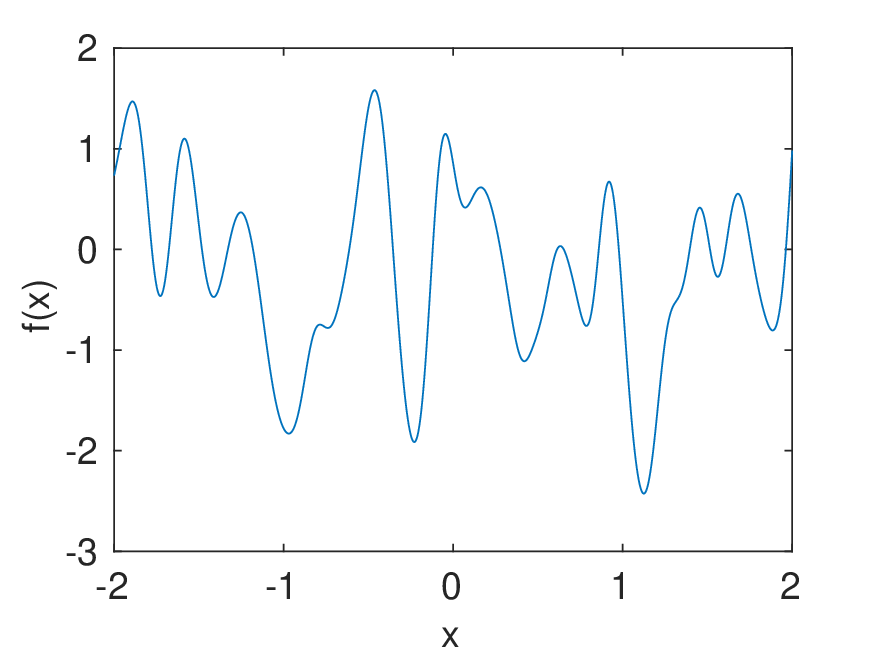}
        \hspace{20pt}
        \includegraphics[width=0.45\textwidth]{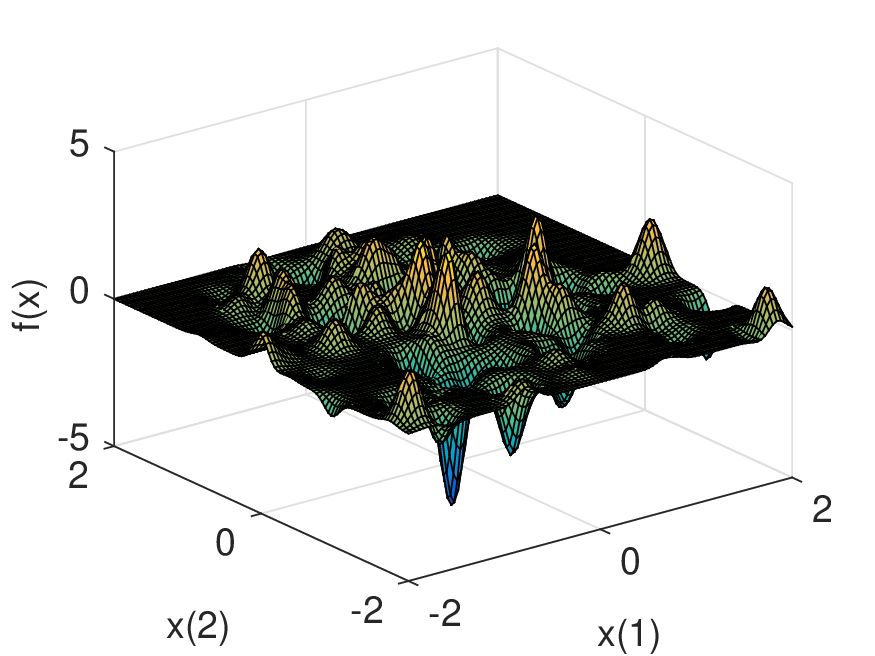}
        \caption{Examples of a function sampled from 1-D GP (left), and a function sampled from 2-D GP (right) with squared exponential kernel and 0 mean functions. These functions can be easier than the ones in Figure~\ref{fig:gpexample} since they have fewer local optima.}%
        \label{fig:esgpexample}
\end{figure}

Since both ES and PES only support squared exponential covariance function and zero mean function in their code right now, and it requires significant changes in their code to accommodate other covariance functions, we created synthetic functions that are different from the ones we used in Section 4 in the paper. The new functions are sampled from 1-D (80 functions) and 2-D GP (20 functions) with squared exponential kernel ($\sigma_f = 0.1$ and $l=1$) and 0 mean. Function examples are shown in Figure~\ref{fig:esgpexample}. 
\begin{figure}
        \centering
        \includegraphics[width=0.45\textwidth]{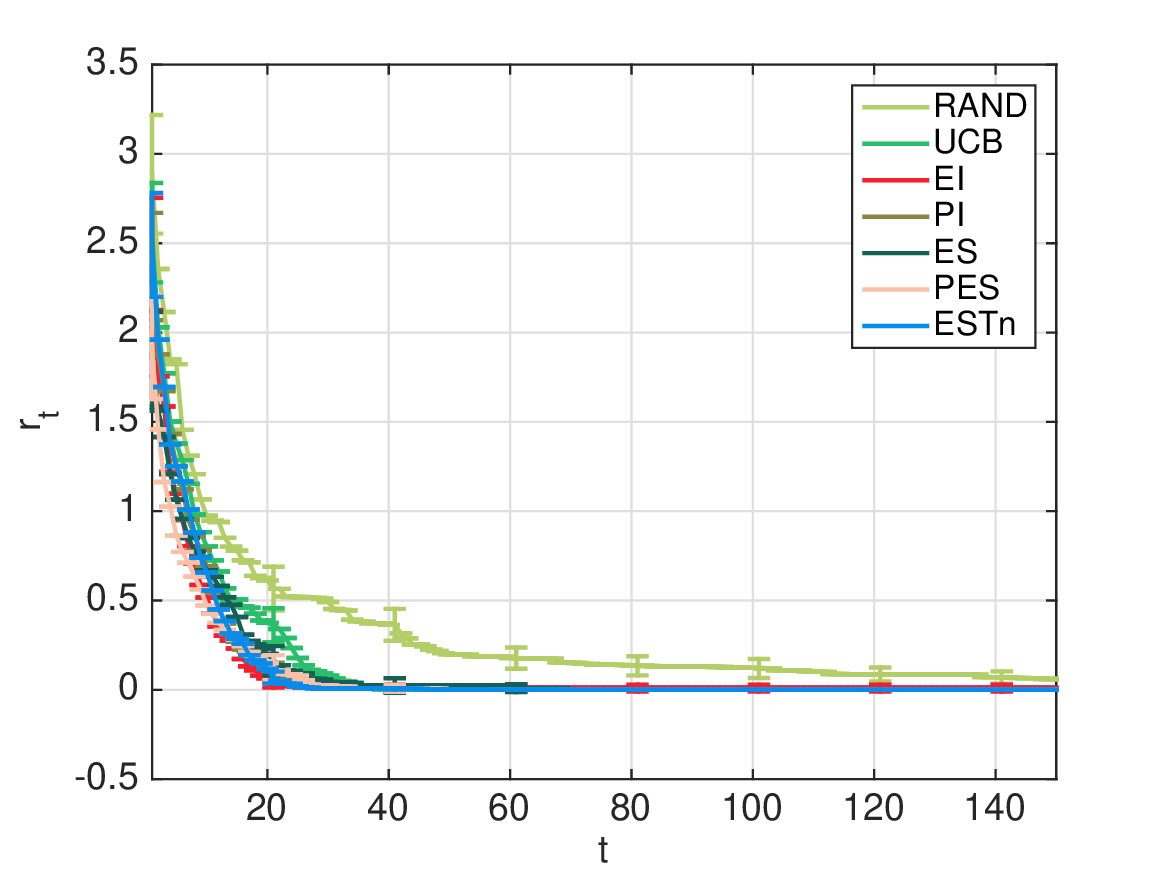}
        \caption{Simple regret %
        for functions sampled from 1-D GP with squared exponential kernel and 0 mean. }%
        \label{fig:es_1d}
\end{figure}

\begin{figure}
        \centering
        \includegraphics[width=0.45\textwidth]{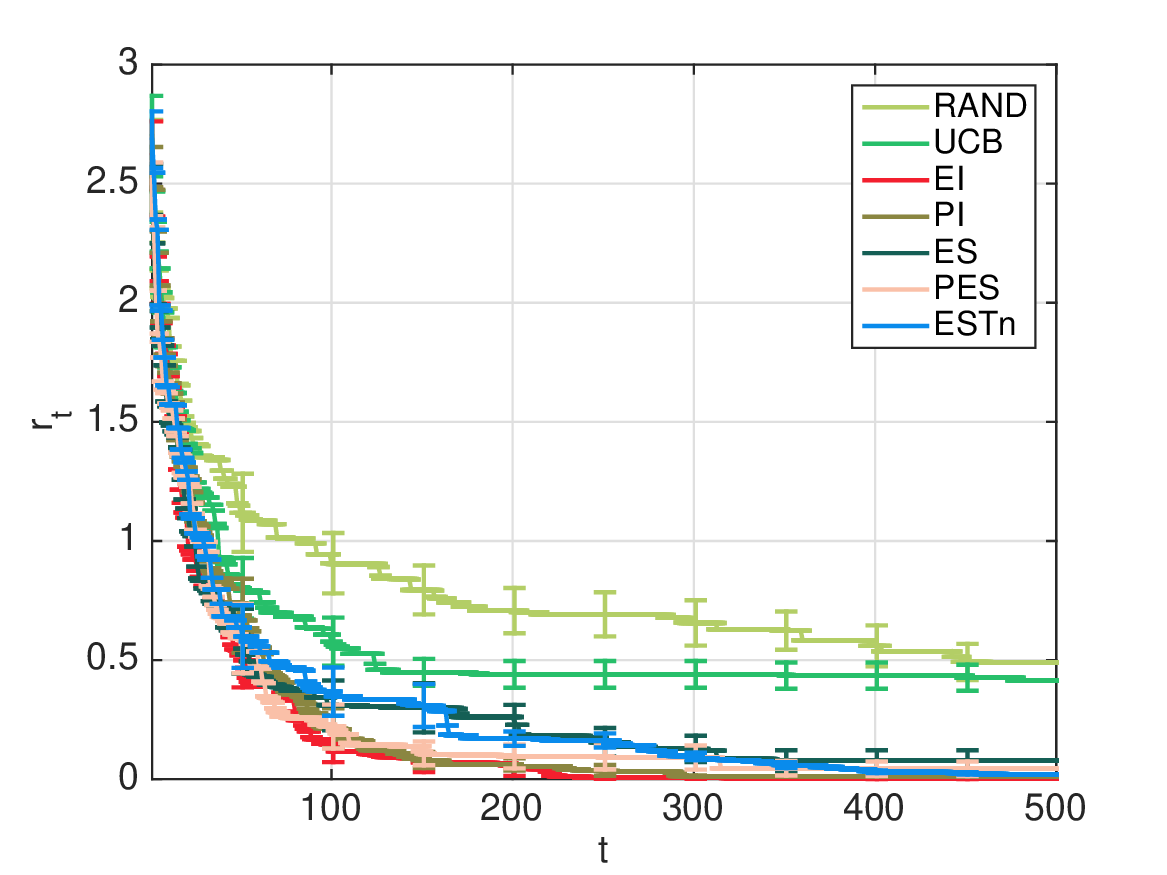}
        \caption{Simple regret  for functions sampled from 2-D GP with squared exponential kernel and 0 mean. }%
        \label{fig:es_2d}
\end{figure}

\begin{figure}
        \centering
        \includegraphics[width=0.45\textwidth]{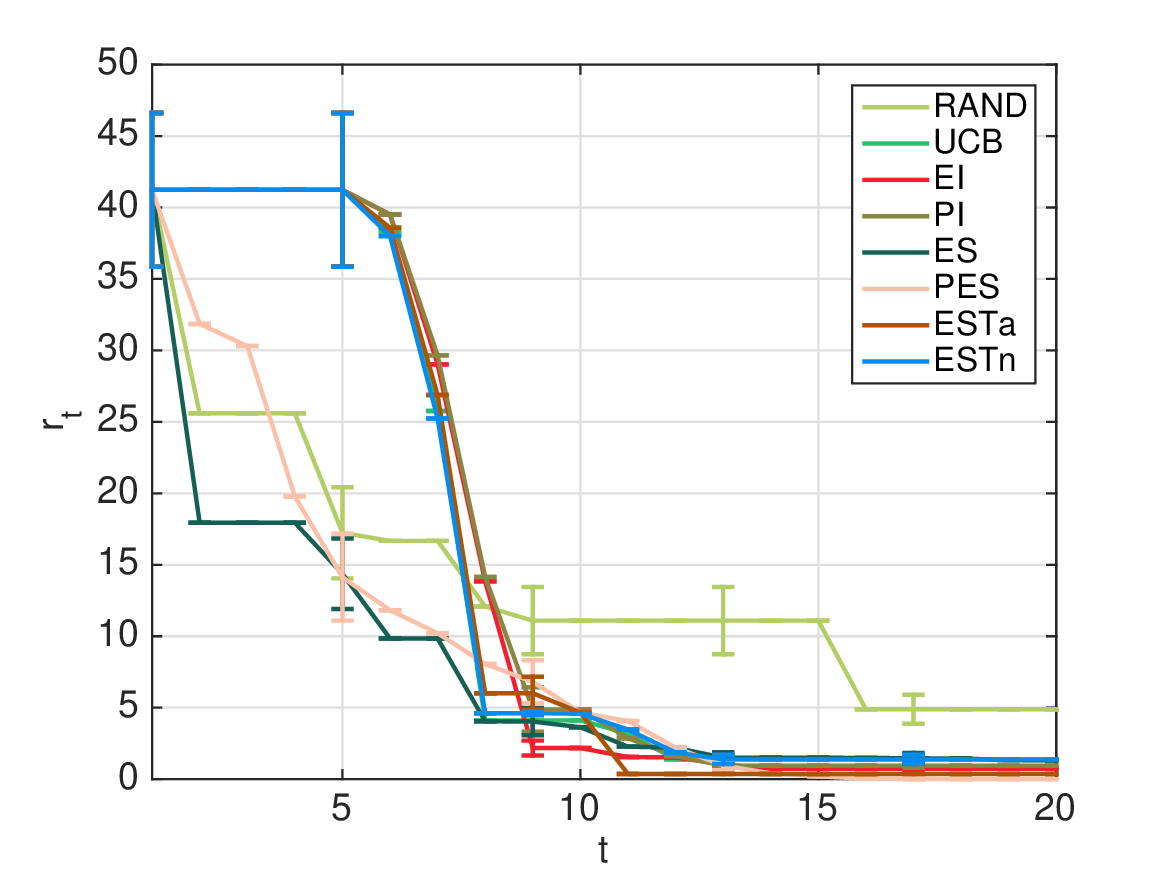}
        \caption{Simple regret for Branin Hoo function. UCB/EI/PI/ESTa/ESTn use the isotropic Mat\'ern kernel with parameters $\ell = 0.1, \sigma_f = 1$; ES/PES use the isotropic squared exponential kernel with $\ell = 0.1, \sigma_f = 1$. }%
        \label{fig:branin}
\end{figure}

We show the results on these synthetic functions in Figure~\ref{fig:es_1d},\ref{fig:es_2d}, and a standard optimization test function, Branin Hoo function, in Figure~\ref{fig:branin}. It is worth noting that ES methods make queries on the most informative points, which are not necessarily the points with low regret. At each round, ES methods make a ``query'' on the black-box function, and then make a ``guess'' of the $\argmax$ of the function (but do not test the ``guess''). %
We plot the regret achieved by the ``guesses'' made by ES methods. For the 1-D GP task, all the methods behave similarly and achieve zero regret except \RND. For the 2-D GP task, EI is the fastest method to converge to zero regret, and in the end ESTn, PI,EI and ES methods achieve similar results. For the test on Branin Hoo function, PES achieves the lowest regret. ESTa converges slightly faster than PES, but to a slightly higher regret.%

We also compared the running time for all the methods in Table~\ref{tb:runtime}. \footnote{All of the methods were run with MATLAB (R2012b), on Intel(R) Xeon(R) CPU E5645  @ 2.40GHz.} 
It is assumed in GP optimization that it is more expensive to evaluate the blackbox function than computing the next query to evaluate using GP optimization techniques. However, in practice, we still want the algorithm to output the next query point as soon as possible. For ES methods, it can be sometimes unacceptable to run them for black-box functions that take minutes to complete a query.

\begin{table}\caption{Comparison on the running time (s) per iteration.}\label{tb:runtime}
\centering
\begin{tabular}{|c|c|c|c|}
\hline
 RAND & UCB & EI & PI \\\hline
0.0002 & 0.075 & 0.079 & 0.076 \\
\hline
\hline
  ESTa & ESTn & ES& PES\\\hline
  0.078 & 0.55 & 56& 106\\
\hline
\end{tabular}

\end{table}

\end{document}